\documentclass[sigconf]{acmart}
\renewcommand\footnotetextcopyrightpermission[1]{}

\makeatletter
\def\@ACM@checkaffil{
    \if@ACM@instpresent\else
    \ClassWarningNoLine{\@classname}{No institution present for an affiliation}%
    \fi
    \if@ACM@citypresent\else
    \ClassWarningNoLine{\@classname}{No city present for an affiliation}%
    \fi
    \if@ACM@countrypresent\else
        \ClassWarningNoLine{\@classname}{No country present for an affiliation}%
    \fi
}
\makeatother

\usepackage{hyperref}
\usepackage{amssymb}
\usepackage{amsfonts}
\usepackage{subfigure}
\usepackage{algorithmic}
\usepackage{graphicx}
\usepackage{xcolor}
\usepackage{makecell}
\usepackage{booktabs}
\usepackage{amsmath}
\usepackage{accents}
\usepackage{statex}
\usepackage[normalem]{ulem}
\usepackage{enumitem}
\usepackage{multirow}
\usepackage[nomargin,inline,marginclue,draft]{fixme}
\usepackage{balance}
\usepackage{changepage}
\usepackage{bm}
\usepackage{setspace}
\usepackage{mathrsfs}
\usepackage{ulem}
\usepackage{verbatim}
\usepackage{diagbox}
\usepackage{pdftexcmds}
\usepackage{catchfile}
\usepackage{ifluatex}
\usepackage{ifplatform}
\usepackage{threeparttable}

\usepackage{comment}

\usepackage[T1]{fontenc}
\usepackage{aecompl}
\usepackage[utf8]{inputenc}
\usepackage{bm} 

\newcommand{\cmark}{\textcolor{blue}{\textbf{$\bm{\checkmark}$}}} 
\newcommand{\xmark}{\textcolor{red}{\textbf{$\bm{\times}$}}} 

\AtBeginDocument{%
  \providecommand\BibTeX{{%
    \normalfont B\kern-0.5em{\scshape i\kern-0.25em b}\kern-0.8em\TeX}}}

\newcommand{\para}[1]{{\vspace{3pt} \bf \noindent #1 \hspace{0pt}}}

\author{Qingyue~Long}
\affiliation{
  \institution{Department of Electronic Engineering\\ BNRist, Tsinghua University}
  \state{Beijing}
  \country{China}
}

\author{Yuan~Yuan}
\affiliation{
  \institution{Department of Electronic Engineering\\ BNRist, Tsinghua University}
  \state{Beijing}
  \country{China}
}

\author{Yong~Li}
\affiliation{
  \institution{Department of Electronic Engineering\\ BNRist, Tsinghua University}
  \state{Beijing}
  \country{China}
}

\copyrightyear{2024}
\acmYear{2024}
\setcopyright{acmlicensed}\acmConference[KDD '25]{Proceedings of the 31th
ACM SIGKDD Conference on Knowledge Discovery and Data Mining}{August
25--29, 2025}{Barcelona, Spain}
\acmBooktitle{Proceedings of the 30th ACM SIGKDD Conference on Knowledge
Discovery and Data Mining (KDD '25), August 25--29, 2025, Barcelona, Spain}

\begin{document}
\title{A Universal Model for Human Mobility Prediction}
\begin{abstract}

Predicting human mobility is crucial for urban planning, traffic control, and emergency response. Mobility behaviors can be categorized into individual and collective, and these behaviors are recorded by diverse mobility data, such as individual trajectory and crowd flow. As different modalities of mobility data, individual trajectory and crowd flow have a close coupling relationship. Crowd flows originate from the bottom-up aggregation of individual trajectories, while the constraints imposed by crowd flows shape these individual trajectories.
Existing mobility prediction methods are limited to single tasks due to modal gaps between individual trajectory and crowd flow.
In this work, we aim to unify mobility prediction to break through the limitations of task-specific models. 
We propose a universal human mobility prediction model (named \textbf{UniMob}), which can be applied to both individual trajectory and crowd flow.
UniMob leverages a multi-view mobility tokenizer that transforms both trajectory and flow data into spatiotemporal tokens, facilitating unified sequential modeling through a diffusion transformer architecture. To bridge the gap between the different characteristics of these two data modalities, we implement a novel bidirectional individual and collective alignment mechanism. This mechanism enables learning common spatiotemporal patterns from different mobility data, facilitating mutual enhancement of both trajectory and flow predictions.
Extensive experiments on real-world datasets validate the superiority of our model over state-of-the-art baselines in trajectory and flow prediction. Especially in noisy and scarce data scenarios, our model achieves the highest performance improvement of more than 14\% and 25\% in MAPE and Accuracy@5. 
\end{abstract}



\maketitle

\section{Introduction}
Human mobility data records the movement of human beings in space over time~\cite{barbosa2018human, long2023practical, yuan2022activity}. It supports various activities~\cite{yuan2024generating, yuan2023learning} and reflects the spatiotemporal dynamics of the city~\cite{rong2024learning, yuan2023spatio}. Consequently, predicting human mobility has significant practical implications, such as urban planning~\cite{zheng2023spatial,zheng2023road}, traffic control~\cite{pappalardo2023future, zeng2024citylight}, and emergency response~\cite{yuan2022activity}. 
Individual trajectory and crowd flow can be treated as two observations of human mobility that represent two different modalities of mobility data. Individual trajectories describe human mobility behavior from a micro perspective, highlighting personal preferences~\cite{zhou2021self}. Conversely, crowd flows encapsulate human movements from a macro perspective, reflecting collective trends~\cite{chang2012dynamic}.
Crowd flows originate from the bottom-up aggregation of individual trajectories, while individual trajectories are influenced by the constraints imposed by crowd flows. This bidirectional influence between individual and collective contributes to the complexity of human mobility.

Many years ago, various models were proposed to model and predict human mobility, such as Lévy flight~\cite{mandelbrot1983fractal}, random walk models~\cite{gonzalez2008understanding}, radiation model~\cite{simini2012universal}, and gravity models~\cite{zipf1946p}. 
Later, trajectory prediction models have been developed to capture individual mobility preferences, such as EPR~\cite{brockmann2006scaling}, MPRW~\cite{yan2017universal}, MobTCast~\cite{xue2021mobtcast} and DeepMove~\cite{feng2018deepmove}. At the same time, flow prediction models like ST-ResNet~\cite{zhang2017deep}, TODE~\cite{zhou2021urban}, DeepCrowd~\cite{jiang2021deepcrowd} and CrowdNet~\cite{cardia2022enhancing} were created to capture the collective movement trends. 
Some research has started integrating trajectory and flow data. For example, GETNext incorporates collective mobility patterns into trajectory prediction~\cite{yang2022getnext}, while TrGNN uses individual mobility data to aid non-recurring flow prediction~\cite{li2021traffic}. 
Although these works have made initial attempts and progress in fusing trajectory and flow data, they remain limited to using other modality data as features. As a result, only a single modality can be predicted, failing to realize the unification of different mobility data modalities.

\begin{figure}[t]
\centering
\includegraphics[width=0.47\textwidth]{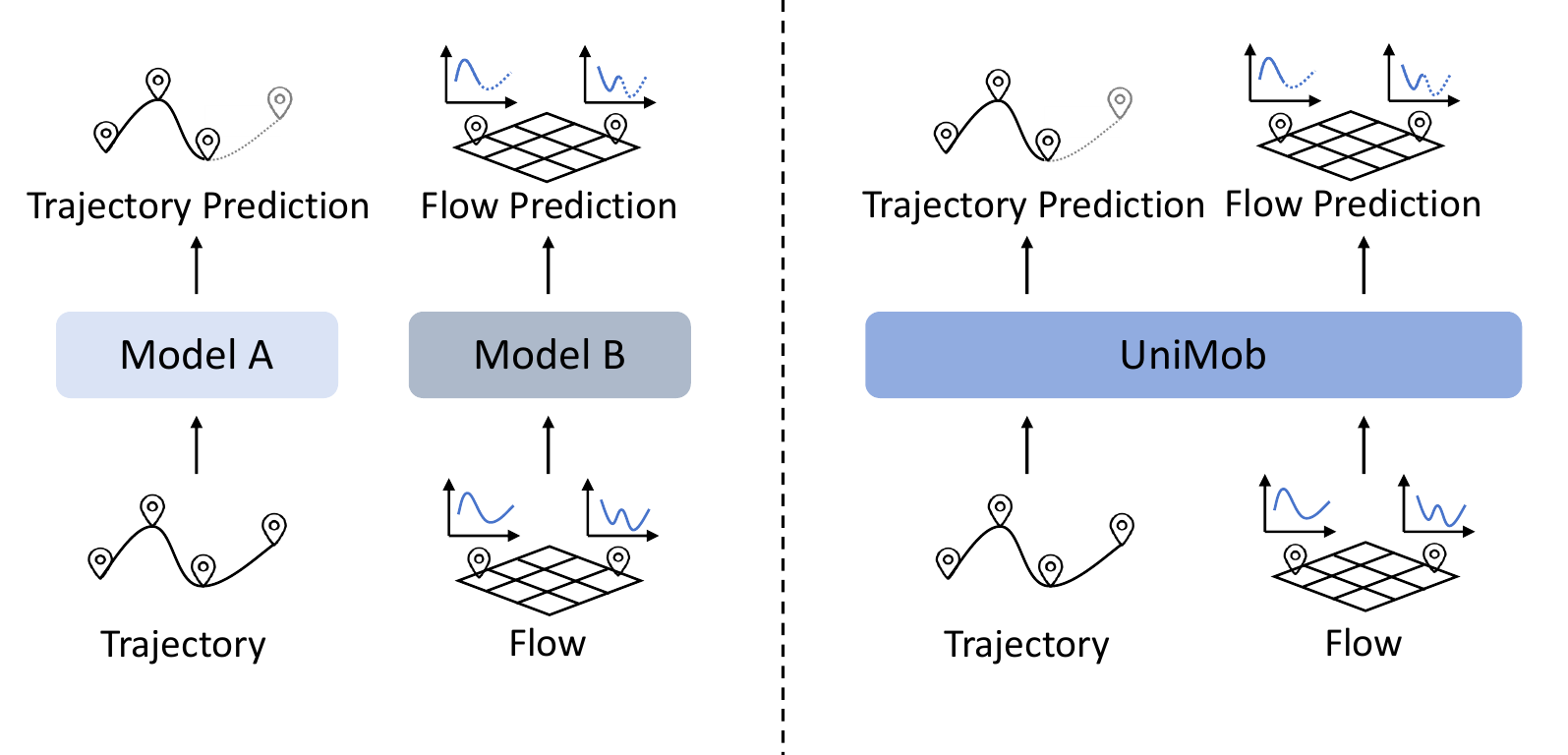}
\vspace{-0.5cm}
\caption{The transition from single model to universal model.}
\label{fig:UniMob}
\vspace{-0.5cm}
\end{figure}

As shown in Figure~\ref{fig:UniMob}, 
we are exploring a natural research question: can we unify human mobility prediction in one universal model? The benefits of such unification are evident: The universal model that learns the common spatiotemporal patterns of two different mobility data in one model can achieve mutual enhancement in trajectory and flow prediction.
However, achieving unified human mobility prediction faces the following critical challenges:
\begin{itemize}[leftmargin=*]
\item \textbf{Diverse data formats of two different mobility data.} The data collection methods for individual trajectories and crowd flows are different, resulting in distinct forms of representation for them. For example, trajectory data records the movement of individuals across different locations over time, while flow data represents the number of people in a specific location that changes over time. These diverse data formats of trajectory and flow make it difficult to represent these data in a unified manner. 

\item \textbf{Significant characteristic differences in two modalities of mobility data.} Trajectory data details individual preferences from a micro perspective, whereas flow data reveals collective trends from a macro perspective. Thus, incorporating these two types of data into a unified training framework and extracting common spatiotemporal patterns from their distinct characteristics is a challenging task.
\end{itemize}

To address these challenges, we propose a universal mobility prediction model, UniMob, that can be applied to both trajectory and flow data.
Firstly, we design a multi-view mobility tokenizer to utilize multiple perspectives of mobility behavior for unified tokenization. Based on sequentially organized trajectory tokens and flow tokens, we implement a diffusion transformer architecture to capture spatiotemporal dynamics inherent in different modalities of mobility data.
Secondly, to address significant characteristic differences in these two modalities, we introduced an innovative bidirectional alignment mechanism that facilitates interaction between trajectories and flows. This mechanism enables the extraction of common spatiotemporal patterns from individual and collective mobility behaviors.
Specifically, the alignment from individual to collective is achieved by aligning aggregated trajectories with flow data, which aids in modeling collective movement trends. 
Conversely, the alignment from collective to individual employs contrastive learning to identify semantically similar flows and trajectories, capturing consistent spatiotemporal patterns at both macro and micro levels.

In this way, UniMob advances towards developing a universal model. UniMob achieves mutual enhancement in trajectory and flow prediction by learning common spatiotemporal patterns from different mobility data. 
Moreover, UniMob has excellent scalability and can flexibly derive into multiple variants according to different requirements, thus adapting to diverse application scenarios. 
Our contributions can be summarized as follows:
\begin{itemize}[leftmargin=*]
\item To our knowledge, we are the first to unify human mobility prediction, exploring the one-for-all model's potential in individual trajectory and crowd flow.

\item We propose a universal mobility prediction model. Multi-view tokenization harmonizes diverse data formats of individual trajectory and crowd flow. Then, the model utilizes bidirectional alignment mechanisms for individual and collective to address the characteristic differences caused by caused by data modalities.

\item Extensive experiments on real-world datasets have validated that UniMob achieves superior performance in trajectory and flow predictions. Further in-depth analysis confirms UniMob's robustness, particularly in handling noisy and scarce data, achieving improvements of over 14\% in MAPE and over 25\% in Accuracy@5. 
\end{itemize}

\section{Related Work}
\paragraph{\textbf{Mobility Prediction}}
Mobility prediction can be divided into individual and collective categories. Individual prediction focuses on personal preferences~\cite{qingyue2024privacy}. For example, Qiao et al.~\cite{gao2019predicting} and Wang et al.~\cite{wang2021attentional} developed a Markov-based model by considering the spatiotemporal characteristics of individual mobility. Collective flow prediction emphasizes modeling collective mobility trends. For example, DeepSTN+~\cite{feng2021context} uses a context-aware spatiotemporal neural network for flow prediction. CrowdNet utilizes graph convolutional networks to achieve flow prediction adapted to various spatial and temporal granularities~\cite{cardia2022enhancing}. 
Researchers have integrated individual and collective mobility data better to understand human mobility~\cite{chen2023multi, bontorin2024mixing,li2023learning}. TrGNN~\cite{li2021traffic} uses vehicle trajectories to infer short-term traffic flow, predicting unseen and non-recurring traffic patterns. GETNext~\cite{yang2022getnext} constructs a global flow graph to integrate transition patterns into trajectory prediction. With the emergence of large language models (LLMs), researchers have begun exploring their potential in mobility prediction, such as LLM-Mob~\cite{wang2023would},  AgentMove~\cite{feng2024agentmove}, TrajAgent~\cite{du2024trajagent} and CoPB~\cite{shao2024beyond}.
However, there is still a gap in using LLM to understand and reason about human behavior. Thus, it is necessary to develop foundational models from scratch, specifically trained on pure mobility data. Table~\ref{tab:model_comparison} compares the advantages of our model with existing solutions. 
In this work, we build a universal model using different types of mobility data, which can effectively predict both trajectories and flows, demonstrating exceptional robustness.

\paragraph{\textbf{Diffusion Models and Foundation Models}}
We focus on human mobility studies based on diffusion models. 
DiffTraj~\cite{zhou2023towards} and ControlTraj~\cite{zhu2024controltraj} combine the generative capabilities of diffusion models with spatiotemporal features derived from trajectories. PriSTI~\cite{liu2023pristi} uses a conditional diffusion framework for spatiotemporal imputation with enhanced prior modeling. TrajGDM~\cite{chu2024simulating} utilizes diffusion models to capture the universal mobility pattern in a trajectory dataset.
Foundation models have revolutionized natural language processing~\cite{achiam2023gpt} and computer vision~\cite{liu2024sora} through their ability to generalize across different applications. 
Building on the success of foundation models in these fields, extending them to the spatiotemporal domain is a natural next step. Although models like UniST~\cite{yuan2024unist} and GPD~\cite{yuan2024spatio}, have made some progress in predicting collective dynamics, there remains a significant gap in universal models capable of adapting to various types of mobility data. 
In this work, we present the UniMob model, marking the first attempt to apply a universal mobility prediction model based on a diffusion transformer.
We provide more discussions of diffusion models in Appendix~\ref{sec::diffusion}.

\begin{table}[t]
    \centering
    \caption{Comparison of UniMob with other mobility prediction models regarding important properties.}
    \vspace{-0.3cm}
    \label{tab:model_comparison}
    {\fontsize{8}{10}\selectfont
    \begin{tabular}{ccccc}
    \toprule
    \makecell{Model} & \makecell{Trajectory\\ Prediction} & \makecell{Flow\\ Prediction} & \makecell{Data \\  Fusion$^{(1)}$} & \makecell{Robustnes$^{(2)}$} \\ 
    \midrule

    DeepMove~\cite{feng2018deepmove} & \cmark & \xmark & \xmark & \xmark \\ 
    SNPM~\cite{yin2023next} & \cmark & \xmark & \xmark & \xmark \\ 
    TrajGDM~\cite{chu2024simulating} & \cmark & \xmark & \xmark & \cmark \\ 
    ST-ResNet~\cite{zhang2017deep} & \xmark & \cmark & \xmark & \xmark \\ 
    STID~\cite{shao2022spatial} & \xmark & \cmark & \xmark & \xmark \\ 
    PriSTI~\cite{liu2023pristi} & \xmark & \cmark & \xmark & \cmark \\ 
    TrGNN~\cite{li2021traffic} & \xmark & \cmark & \cmark & \xmark \\ 
    GETNext~\cite{yang2022getnext} & \cmark & \xmark & \cmark & \xmark \\ 
    LLM-Mob~\cite{wang2023would} & \cmark & \xmark & \xmark & \cmark \\ \midrule
    UniMob & \cmark & \cmark & \cmark & \cmark \\ 
    \bottomrule
    \end{tabular}

    \vspace{2mm} 

    \footnotesize
    \begin{tabular}{ll}
        (1) & Use of multi-source data (trajectory and flow). \\
        (2) & Keep ability with noisy or scarce data. \\
    \end{tabular}}
\vspace{-0.5cm}
\end{table}

\section{PRELIMINARIES}
\begin{figure*}[t]
\centering
\includegraphics[width=0.95\textwidth]{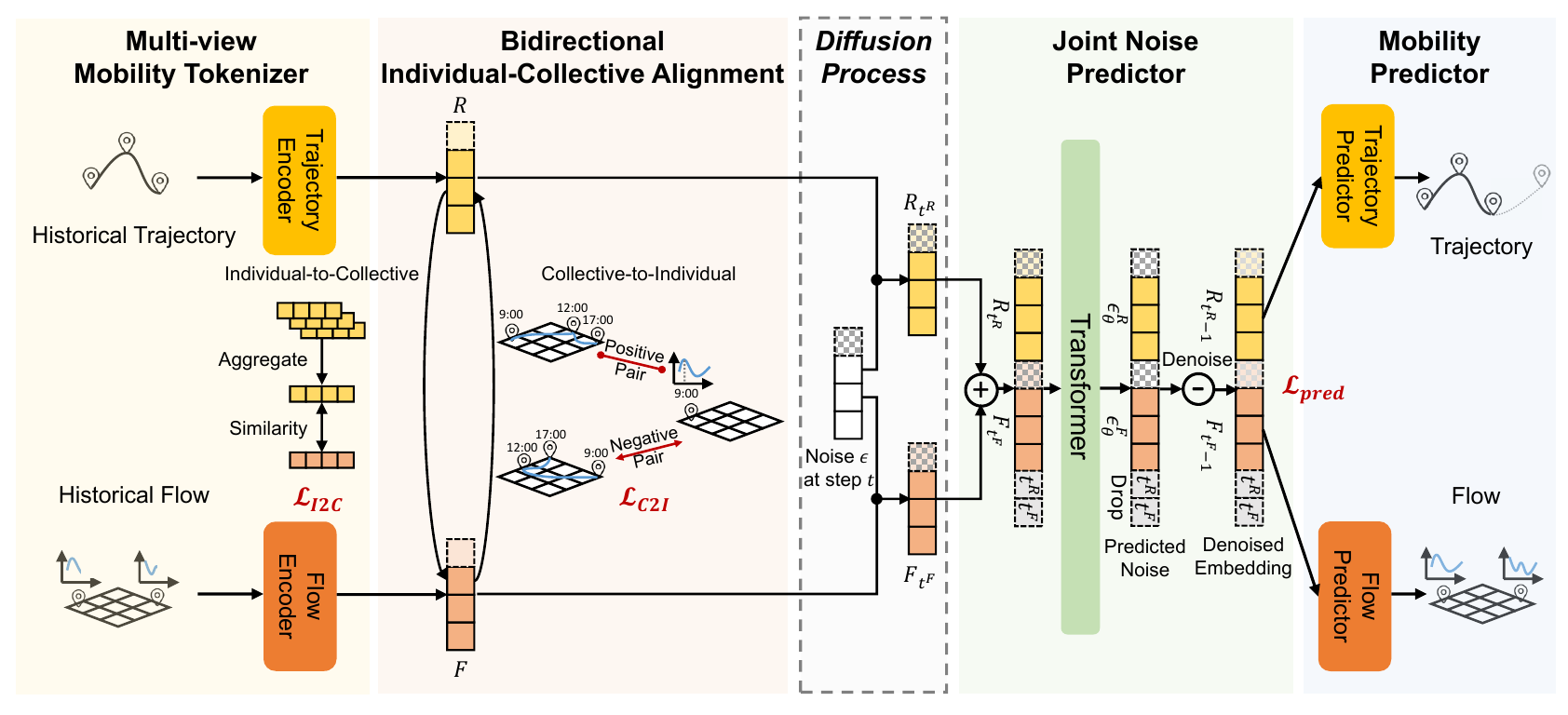}
\vspace{-0.3cm}
\caption{The overview architecture of UniMob, which consists of four modules: (1) Multi-view Mobility Tokenizer, (2) Bidirectional Individual-Collective Alignment, (3) Joint Noise Predictor, (3) Mobility Predictor.} \label{fig:framework}
\end{figure*} 

\subsection{Problem Definition}
Human mobility data can be divided into two types: individual and collective. Trajectories can describe individual mobility, while flow data can characterize collective mobility.

\para{Definition 1: (Individual Trajectory).} An individual trajectory can be defined as $X^{traj}=\{(l_1, t_1),(l_2, t_2),...,(l_n, t_n)\}$, where each location $l_i$ is represented as the form in latitude and longitude coordinates or a region ID.

\para{Definition 2: (Crowd Flow).} Crowd flow includes inflow and outflow, defined as the number of people entering or leaving a region within a given time interval. The crowd flow for a region $l$ can be represented as $X^{flow}_l \in \mathbb{R}^{N \times T}$, where $T$ is the number of
time intervals, and $N$ is the dimension of the flow, such as $N = 2$ for inflow and outflow. The entire city's flow can be represented as $Y \in \mathbb{R}^{N \times T \times L}$, where $L$ is the number of regions.

\para{Problem Statement: (Mobility Prediction).} Given $p$ historical records of mobility data (which can be trajectory $X^{traj}_{[t-p:t]}$ or flow $X^{flow}_{[t-p:t]}$), our goal is to predict the future $k$ steps $X^{traj}_{[t:t+k]}$ or $X^{flow}_{[t:t+k]}$.

\subsection{Denoising Diffusion Probabilistic Model}
Diffusion models use latent variable models, denoted as $p_\theta(x_0) :=\int p_\theta(x_{0:T}) dx_{1:T}$. The latent variables $x_1, ..., x_T$ have the same dimension as the data $x_0 \sim q(x_0)$. The model uses two Markov chains: a forward chain that perturbs data into noise and a reverse chain that converts noise back into data.
The forward diffusion process:
\begin{equation}\label{equ:DDPM1}
q\left(\mathbf{x}_{1: T} \mid \mathbf{x}_{0}\right):=\prod_{t=1}^{T} q\left(\mathbf{x}_{t} \mid \mathbf{x}_{t-1}\right),
\end{equation}
where $q\left(\mathbf{x}_{t} \mid \mathbf{x}_{t-1}\right):=\mathcal{N}\left(\sqrt{1-\beta_{t}} \mathbf{x}_{t-1}, \beta_{t} \mathbf{I}\right)$. Equivalently, $x_t$ can be expressed as $x_{t}=\sqrt{\alpha_{t}} x_{0}+\left(1-\alpha_{t}\right) \epsilon$ for $\epsilon \sim \mathcal{N}(0, \mathbf{I})$, with $\alpha_{t}=\sum_{i=1}^{t}\left(1-\beta_{i}\right)$.

The reverse process denoises $x_t$ to retrieve $x_0$, where $\mathbf{x}_{T} \sim \mathcal{N}(\mathbf{0}, \mathbf{I})$. Assuming $p_\theta(x_{t-1}|x_t)$ follows a normal distribution:
\begin{equation}\label{equ:DDPM2}
\left\{
\begin{array}{l}
p_{\theta}\left(\mathbf{x}_{0: T}\right) := p\left(\mathbf{x}_{T}\right) \prod_{t=1}^{T} p_{\theta}\left(\mathbf{x}_{t-1} \mid \mathbf{x}_{t}\right), \\
p_{\theta}\left(\mathbf{x}_{t-1} \mid \mathbf{x}_{t}\right) := \mathcal{N}\left(\mathbf{x}_{t-1}; \boldsymbol{\mu}_{\theta}\left(\mathbf{x}_{t}, t\right), \sigma_{\theta}\left(\mathbf{x}_{t}, t\right) \mathbf{I}\right)
\end{array}
\right.
\end{equation}

Ho et al.~\cite{ho2020denoising} introduced denoising diffusion probabilistic models:
\begin{equation}\label{equ:DDPM4}
\begin{cases}
\boldsymbol{\mu}_{\theta}\left(\mathbf{x}_{t}, t\right)=\frac{1}{\alpha_{t}}\left(\mathbf{x}_{t}-\frac{\beta_{t}}{\sqrt{1-\alpha_{t}}} \boldsymbol{\epsilon}_{\theta}\left(\mathbf{x}_{t}, t\right)\right),\\
\sigma_{\theta}\left(\mathbf{x}_{t}, t\right)=\tilde{\beta}_{t}^{1 / 2}, \tilde{\beta}_{t}=\left\{\begin{array}{ll}
\frac{1-\alpha_{t-1}}{1-\alpha_{t}} \beta_{t} & t>1 \\
\beta_{1} & t=1
\end{array}\right.
\end{cases}
\end{equation}
where $\epsilon_\theta$ is a trainable denoising function. The objective for training the reverse process is:
\begin{equation}\label{equ:DDPM5}
\min _{\theta} \mathcal{L}(\theta):=\min _{\theta} \mathbb{E}_{\mathbf{x}_{0} \sim q\left(\mathbf{x}_{0}\right), \boldsymbol{\epsilon} \sim \mathcal{N}(\mathbf{0}, \mathbf{I}), t}\parallel\boldsymbol{\epsilon}-\boldsymbol{\epsilon}_{\theta}\left(\mathbf{x}_{t}, t\right)\parallel_{2}^{2},
\end{equation}
where $\mathbf{x}_{t}=\sqrt{\alpha_{t}} \mathbf{x}_{0}+\left(1-\alpha_{t}\right) \boldsymbol{\epsilon}$. This can be seen as a weighted variational constraint on the negative log-likelihood, reducing the significance of terms at low $t$ when little noise is present.

\section{Method}
To unify diverse data formats and varied data characteristics, our universal solution includes two key designs:
1) The mobility tokenizer transforms trajectory and flow into a unified spatiotemporal token format, facilitating the utilization of the powerful diffusion transformer architecture.
2) The individual and collective alignment is designed to jointly train different data types within the same model framework, effectively aligning the differences in individual and collective mobility behavior. 

\subsection{Overall framework}
The framework of UniMob is shown in Figure~\ref{fig:framework}, which consists of four modules:
\begin{itemize}[leftmargin=*]
    \item \textbf{Multi-view Mobility Tokenizer:} It first standardizes different data types through tokenization and then uses two mobility encoders to capture the spatiotemporal dynamics of trajectory and flow from multiple views.
    \item \textbf{Bidirectional Individual-Collective Alignment:} This module consists of two parts: Individual-to-Collective alignment and Collective-to-Individual alignment. Individual-to-Collective alignment utilizes the I2C loss function, which captures collective dynamics by aligning aggregated trajectories with the flow. Collective-to-Individual alignment uses the C2I loss function, which employs contrast learning to align individual trajectory and crowd flow with the same spatiotemporal pattern.
    
    \item \textbf{Joint Noise Predictor:} Learning the spatiotemporal distribution of mobility behavior can be represented as a denoising diffusion process, utilizing a joint noise predictor for trajectories and flows to model individual and collective mobility effectively.
    
    \item \textbf{Mobility Predictor:} The goal of the mobility predictor is to decode high-dimensional spatiotemporal features, capture the dynamic changes in trajectory or flow, and achieve accurate prediction results.
\end{itemize}
Overall, the training of the model is supervised by three types of losses: I2C loss, C2I loss, and prediction loss. The first two losses are designed to promote alignment between individuals and collective, while the prediction loss targets the joint noise predictor of the diffusion model to estimate the noise to be removed and improve prediction accuracy. The combined effect of these losses significantly improves the accuracy and robustness of UniMob in understanding and predicting complex mobility behaviors.

\subsection{Multi-view Mobility Tokenizer}
To address the challenge of diverse data formats, tokenization is applied to process different data types into a unified sequential format. Subsequently, two encoders capture the spatiotemporal dynamics inherent in the mobility data from multiple views.

\subsubsection{\textbf{Tokenization}}
Tokenization is applied to mobility data such as trajectories and flow data, converting it into compact token sequences to enable more efficient computation and standardized processing.
We divide the trajectory $X^{traj}$ and flow $X^{flow}$ into several continuous overlapping or non-overlapping tokens, each with a length of $p$. Therefore, the total number of input tokens is $C=\frac{(T-p)}{Q}$, where $T$ denotes the total number of time intervals and $Q$ represents the horizontal sliding stride.

\subsubsection{\textbf{Trajectory Encoder}}\label{sec:encoder}
In this module, we model trajectory from the spatial and temporal perspectives. 
Firstly, to capture the geographical continuity of mobility, we construct a spatial graph \( G = (V, E) \). The nodes \( V \) represent all visited locations, and the edges \( E \) define the connections between these locations. Each edge \( e = (u, v) \) is an unordered pair with a positive weight \( w_{uv} \), representing the Euclidean distance between locations \( u \) and \( v \).
We use a graph embedding approach $\mathcal{F}(\cdot)$ to get the location embedding $S$, which is denoted as follows:
\begin{equation}\label{equ:graph}
S={\mathcal{F}}_\theta(G),
\end{equation}
We obtain a spatial embedding matrix $S \in \mathbb{R}^{L \times W}$, where $L$ is the number of regions, and $W$ is the dimension of embedding.

Then, to capture the periodicity of time, we use temporal embedding matrices $H \in \mathbb{R}^{T_h \times W}$ and $D \in \mathbb{R}^{T_d \times w}$ to represent the time features. \(T_h\) is the number of time slots in a day (determined by the sampling frequency), and \(T_d = 7\) is the number of days in a week.

Finally, the trajectory embedding $R$ is obtained by concatenating the spatial and temporal embeddings, represented as:
\begin{equation}\label{equ:traj_embedding}
R_t=[S_t; H_t; D_t],
\end{equation}
where \(R_t\) is the vector representation of the \(t\)-th trajectory point.

\subsubsection{\textbf{Flow Encoder}}
The flow can obtain corresponding spatial and temporal embeddings like the trajectory. Moreover, the flow has an extra embedding for historical values, achieved by mapping the raw historical time series $X^{flow}_{[t-p:t]}$ into the latent space $V_t \in \mathbb{R}^{W}$:
\begin{equation}\label{equ:flow_embedding}
V_t = FC(X^{flow}_{[t-p:t]}),
\end{equation}
where $FC(\cdot)$ is a fully connected layer and $W$ is the dimension of embedding.

Thus, we connect the spatial embedding, the temporal embedding, and historical embedding to obtain the flow embedding $F$:
\begin{equation}\label{equ:flow_embedding}
F_t=[S_t; H_t; D_t; V_t].
\end{equation}

\subsection{Joint Noise Predictor}
Formally, suppose we have two types of mobility data sampled from the distribution $q(R_0, F_0)$. Our goal is to design a diffusion-based model that can capture some related distributions determined by $q(R_0, F_0)$, specifically the marginal distributions $q(R_0)$ and $q(F_0)$.

According to Bao et al.~\cite{bao2023one}, different forms of distributions can be unified into the general form of $\mathbb{E}[\epsilon^R, \epsilon^F \mid R_{t^R}, F_{t^F}]$, where $t^R$ and $t^F$ are two different timesteps, and $R_{t^R}$ and $F_{t^F}$ are the corresponding perturbed data. Particularly, a maximum timestep $T$ means marginalizing it. By setting $t^F = T$, we have $\mathbb{E}[\epsilon^R \mid R_{t^R}, F_T] \approx \mathbb{E}[\epsilon^R \mid R_{t^R}]$, which corresponds to the marginal distribution $q(R_0)$. 
Similarly, by setting $t^R = T$, we have $\mathbb{E}[\epsilon^F \mid F_{t^F}, R_T] \approx \mathbb{E}[\epsilon^F \mid F_{t^F}]$, which corresponds to the marginal distribution $q(F_0)$.

Inspired by the unified view, we learn $\mathbb{E}[\epsilon^R, \epsilon^F \mid R_{t^R}, F_{t^F}]$ for all $0 \leq t^R, t^F \leq T$ to model all relevant distributions determined by $q(R_0, F_0)$. We employ a joint noise prediction network $\epsilon_\theta(R_{t^R}, F_{t^F}, t^R, t^F)$ to predict the noise injected into $R_{t^R}$ and $F_{t^F}$ together:
\begin{equation}\label{equ:predicted noise}
\hat{\epsilon}_{\theta}=\epsilon_\theta(R_{t^R}, F_{t^F}, t^R, t^F),
\end{equation}

We train a joint noise prediction network based on the trajectory and flow embeddings obtained in Section~\ref{sec:encoder}. Naturally, to capture the spatiotemporal correlations of mobility data, we use a transformer-based backbone in UniMob to process inputs from different mobility data types.

\subsection{Mobility Predictor}
After the denoising process in the latent space is complete, the target of the predictor is to transform the embeddings back into the form of the original data.

\subsubsection{\textbf{Trajectory Predictor}}
For the discrete space domain, we add an inverse step at the end of the denoising process, which converts the real-valued embedding of $\hat{R}_0$ to the probability distribution of locations $d=\{d_1,d_2,...,d_N\}$ as follows:
\begin{equation}
P(E) = (E^T E)^{-1} E^T,
\end{equation}
where $E$ denotes Embedded Matrix, $E^T$ denotes the transpose of matrix $E$, and $(\cdot)^{-1}$ represents the matrix inverse.

\begin{equation}
d = \hat{R}_0*P(E),
\end{equation}
where $d$ denotes the probability of visiting each location in the trajectory.

\begin{equation}
i = argmax(d),
\end{equation}
where $argmax(\cdot)$ function is used to find the maximum value index in a vector, and $i$ denotes the next location id.

\subsubsection{\textbf{Flow Predictor}}
The regression layer makes predictions based on the following:
\begin{equation}\label{equ:flow_embedding}
Y_{[t:t+f]} = FC(\hat{F}_0),
\end{equation}
where $FC(\cdot)$ is a fully connected layer and $Y_{[t:t+f]}$ is future flow.

\subsection{Bidirectional Individual-Collective Alignment}
To address significant characteristic differences in two modalities of mobility data, we design several alignment mechanisms, including Individual-to-Collective and Collective-to-Individual alignment. These methods enable collaborative perception and complementary expression of multi-source mobility data. 
After obtaining well-interacted representations of trajectories and flows, we employ the prediction loss from a joint noise prediction network.

\subsubsection{\textbf{Individual-to-Collective Alignment}}
The alignment from individuals to collective is designed to aggregate individual movements from the bottom up, collaborating with collective mobility patterns to capture collective dynamics. Thus, we utilize I2C (Individual to Collective) loss to align trajectory with flow. Specifically, we first aggregate multiple trajectory embeddings:
\begin{equation}\label{equ:aggregate}
R^{all} = R^{u_1} + R^{u_2} + ... + R^{u_n},
\end{equation}
where \(R^{all}\) is the aggregated trajectory embeddings of multiple users, while \(R^{u_n}\) represents the trajectory embedding of user \(u_n\).

This aggregated embedding captures and displays the collective behavioral characteristics and trends. Then, we interact this aggregated result with the flow embedding. We can optimize their collaborative representation by maximizing the similarity between these two embeddings:
\begin{equation}\label{equ:flow_loss}
\mathcal{L}_{I2C} = 1 - \frac{{u} \cdot {v}}{||u|| ||v||},
\end{equation}
Through the I2C loss, we can more accurately capture and describe the dynamic changes of collective in space and time.

\subsubsection{\textbf{Collective-to-Individual Alignment}}
The alignment from the collective to the individual aims to impose constraints on individual behavior through collective mobility patterns, better modeling individual movement preferences.
Through contrastive learning, we utilize C2I (Collective to Individual) loss to identify common spatiotemporal patterns in micro-level trajectories and macro-level flows.
We determine positive and negative samples based on the spatiotemporal consistency between trajectory data and flow data:
\begin{itemize}[leftmargin=*]
    \item \textbf{Positive samples $R^+$:} If at a specific time and location, the flow data shows a peak (for example, a traffic hub during the morning rush hour), then individual trajectory data that matches this time and location are considered positive samples. This implies that these individual trajectories are aligned with the spatiotemporal patterns of the flow data.
    \item \textbf{Negative samples $R^-$:} Trajectories that appear at the same location when there is no flow peak or at different times and locations are considered negative samples. These trajectories do not align with the spatiotemporal patterns of the flow data.
\end{itemize}

For example, in a commercial area, we observe that the flow of people peaks between 5 PM and 6 PM every day, mainly because most office workers leave work during this period. The positive samples are the trajectories recorded in the commercial area during this time. These trajectories reflect the common behavior of most office workers leaving work on time, showing the main flow direction in the commercial area. On the other hand, negative samples are the trajectories recorded in other areas (such as suburbs or residential areas) during the same period. These represent people who leave work early or those whose workplaces are not in the commercial area.
By comparing these samples, we can align collective mobility dynamics with individual mobility preferences, ensuring consistency between micro-level and macro-level mobility patterns.

According to the contrast learning framework, we maximize the similarity between anchors and augmented positive examples while minimizing the similarity between anchors and negative examples lost through InfoNCE~\cite{oord2018representation}:
\begin{equation}\label{equ:flow_loss}
\mathcal{L}_{C2I} =\sum_{i=1}^{N} \log \frac{\boldsymbol{\varphi}\left(\boldsymbol{F}, {\boldsymbol{R}}^{+}\right)}{\sum_{\boldsymbol{R}^{-} \in S} \boldsymbol{\varphi}\left(\boldsymbol{F}, \boldsymbol{R}^{-}\right)},
\end{equation}
where $\boldsymbol{\varphi}(i, j)=\exp (\operatorname{sim}(i, j) / \tau)$ measures the correlation between two representations, where $\operatorname{sim}(\cdot, \cdot)$ denotes the cosine similarity function, ${R}^{+}$ represents the positive sample. $S$ is a set of random negative samples from the same batch.

\subsection{Training}
We model the embedding of trajectories and flows using a joint noise prediction network with the following prediction loss function:
\begin{equation}\label{equ:prediction_loss}
\mathcal{L}_{pred} = {\mathbb{E}_{\boldsymbol{R}_{0}, \boldsymbol{F}_{0}, \boldsymbol{\epsilon}^{R}, \boldsymbol{\epsilon}^{F}, t^{R}, t^{F}}||\boldsymbol{\epsilon}_{\boldsymbol{\theta}}(\boldsymbol{R}_{t^{R}}, \boldsymbol{F}_{t^{F}}, t^{R}, t^{F})-[\boldsymbol{\epsilon}^{R}, \boldsymbol{\epsilon}^{F}]||_{2}^{2},},
\end{equation}
where $[, ]$ denotes concatenation, ${\epsilon}^{R}$ and ${\epsilon}^{F}$ are sampled from standard Gaussian distributions, and $t_R$ and $t_F$ are uniformly sampled from $\{1, 2, . . . , T\}$ independently.

Finally, the total loss is expressed as:
\begin{equation}\label{equ:loss}
\mathcal{L}_{total} = \alpha\mathcal{L}_{I2C} + \beta\mathcal{L}_{C2I} + \gamma\mathcal{L}_{pred},
\end{equation}
The weights $\alpha$, $\beta$, and $\gamma$ correspond to the three loss terms, allowing the total loss function $\mathcal{L}_{total}$ to simultaneously enable trajectory and flow interactions and accurately predict mobility.

\subsection{Variants}
\begin{figure}[t]
\centering
\includegraphics[width=1.0\linewidth]{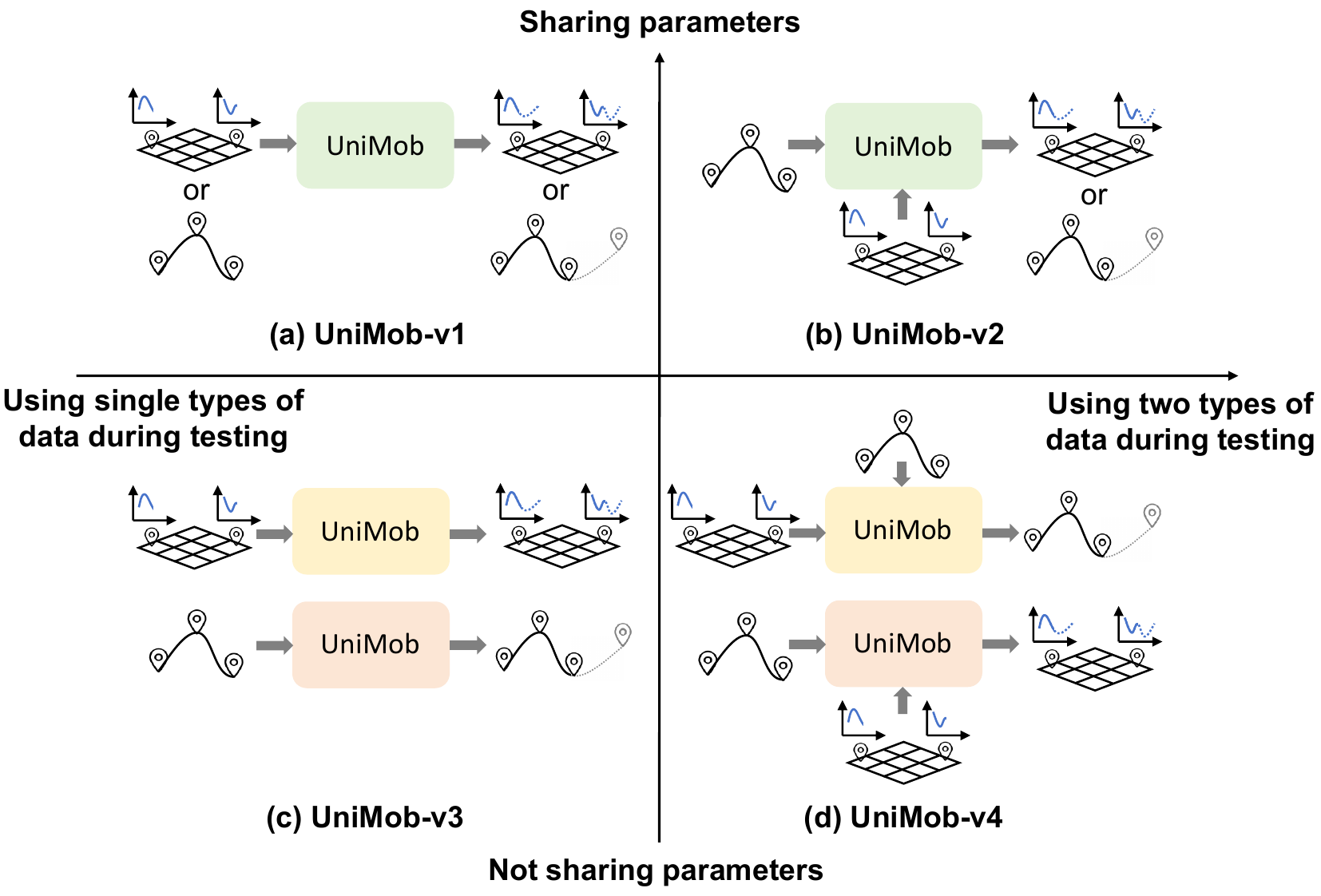}
\vspace{-0.3cm}
\caption{Four model variants based on whether parameters are shared and whether two types of mobility data are used during testing.}
\label{fig:use}
\vspace{-0.3cm}
\end{figure}

In the real world, universal models have a wide range of application scenarios and diverse application requirements. For example, in some cases, there is only one type of mobility data available; in other cases, to save computational resources, it is desirable to use a shared set of parameters to perform flow and trajectory predictions simultaneously. Therefore, we design four model variants to ensure our model can flexibly adapt to different application scenarios and possess greater practicality. As shown in Figure~\ref{fig:use}, these variants are based on whether parameters are shared and whether both types of mobility data are used during testing:
\begin{itemize}[leftmargin=*]
\item \textbf{UniMob-v1}: This variant is designed for scenarios with limited data types and constrained computational resources.
\item \textbf{UniMob-v2}: This variant is suitable for scenarios with multiple types of mobility data but needs to save computational resources.
\item \textbf{UniMob-v3}: This variant can handle scenarios with rich data types but limited computational resources.
\item \textbf{UniMob-v4}: This variant is designed for users with rich data types and ample computational resources.
\end{itemize}

\section{experiments}
\subsection{Experimental Settings}
\subsubsection{Dataset}
We conduct extensive experiments on three real-world mobility datasets from Shanghai, Senegal, and Xinjiang. Each dataset includes both trajectory and flow data. The details of datasets are summarized in Appendix~\ref{sec::datasets_info}. The experiment section only presents the results for the Shanghai and Senegal datasets. Detailed results for the Xinjiang dataset can be found in Appendix~\ref{sec::Results}.

\subsubsection{Baselines}

We compare the performance of our model with state-of-the-art baselines. Previous methods could only accomplish one type of mobility data prediction task, so the baseline methods are divided into trajectory and flow prediction. 
For \textit{Flow Prediction}, we compared our model with six SOTA baselines (\textbf{HA}~\cite{sun2020predicting}, \textbf{VAR}~\cite{lu2016integrating}, \textbf{ST-ResNet}~\cite{zhang2017deep}, \textbf{MSDR}~\cite{liu2022msdr}, \textbf{STID}~\cite{shao2022spatial}, and \textbf{PriSTI}~\cite{liu2023pristi}.
For \textit{Trajectory Prediction}, we compared our model with seven SOTA baselines (\textbf{Markov}~\cite{gambs2012next}, \textbf{LSTM}~\cite{Kong2018HST}, \textbf{DeepMove}~\cite{feng2018deepmove}, \textbf{STAN}~\cite{luo2021stan}, \textbf{SNPM}~\cite{yin2023next},
\textbf{TrajGDM}~\cite{chu2024simulating}, and \textbf{GETNext}~\cite{yang2022getnext}.
We provide the details of baselines in Appendix~\ref{sec::baselines}.

\subsubsection{Metrics}
For trajectory prediction, we use \textit{Accuracy@k} to sort candidate locations by model-predict probabilities and check if the true position falls within the top k predictions.
For flow prediction, we choose mean absolute errors (\textit{MAE}), Mean Absolute Percentage Error (\textit{MAPE}), and root mean squared errors (\textit{RMSE}) as the evaluation metrics. \textit{MAE} is the mean absolute error between predicted and ground truth values. \textit{MAPE} is the mean absolute percentage error between the predicted and ground truth values. \textit{RMSE} is the square root of the mean squared error between the predicted and ground truth values. 

\begin{table*}[t]
\centering
\caption{Overall Performance on Shanghai and Senegal datasets.}
\vspace{-0.3cm}
\scalebox{0.78}{ 
\begin{tabular}{lcccccccccccc}
\toprule
& \multicolumn{6}{c}{\textbf{Shanghai Dataset}} & \multicolumn{6}{c}{\textbf{Senegal Dataset}} \\
\cmidrule(lr){2-7} \cmidrule(lr){8-13}
& \multicolumn{3}{c}{\textbf{Flow Prediction}} & \multicolumn{3}{c}{\textbf{Trajectory Prediction}} & \multicolumn{3}{c}{\textbf{Flow Prediction}} & \multicolumn{3}{c}{\textbf{Trajectory Prediction}}\\
\cmidrule(lr){2-4} \cmidrule(lr){5-7} \cmidrule(lr){8-10} \cmidrule(lr){11-13}
\textbf{} & \textbf{MAE} & \textbf{MAPE(\%)} & \textbf{RMSE} & \textbf{Acc@1} & \textbf{Acc@3} & \textbf{Acc@5} & \textbf{MAE} & \textbf{MAPE(\%)} & \textbf{RMSE} & \textbf{Acc@1} & \textbf{Acc@3} & \textbf{Acc@5} \\
\midrule
HA & 35.19 & 29.76 & 42.72 & - & - & - & 20.75 & 19.32 & 31.17 & - & - & - \\
VAR & 28.25 & 25.61 & 40.14 & - & - & - & 17.20 & 15.95 & 28.43 & - & - & - \\
ST-ResNet & 21.54 & 19.02 & 34.18 & - & - & - & 15.95 & 13.84 & 26.37 & - & - & - \\
MSDR & 20.01 & 17.84 & 32.63 & - & - & - & 14.08 & 13.26 & 25.04 & - & - & - \\
STID & 18.72 & 15.17 & 30.40 & - & - & - & 13.52 & 12.31 & 23.19 & - & - & - \\
PriSTI & \underline{18.40} & \underline{14.59} & \underline{29.71} & - & - & - & \underline{13.28} & \underline{12.15} & \underline{22.80} & - & - & - \\
Markov & - & - & - & 0.2825 & 0.3986 & 0.5012 & - & - & - & 0.3894 & 0.4418 & 0.5828 \\
LSTM & - & - & - & 0.3401 & 0.4298 & 0.5737 & - & - & - & 0.4573 & 0.5185 & 0.6509 \\
DeepMove & - & - & - & 0.3813 & 0.4672 & 0.6191 & - & - & - & 0.4980 & 0.5764 & 0.7125 \\
STAN & - & - & - & 0.3975 & 0.4746 & 0.6303 & - & - & - & 0.5105 & 0.6042 & 0.7303 \\
SNPM & - & - & - & 0.4012 & 0.4797 & 0.6378 & - & - & - & 0.5236 & 0.6260 & 0.7591 \\
GETNext  & - & - & - & 0.4063 & 0.4836 & 0.6415 & - & - & - & 0.5251 & 0.6287 & 0.7638 \\
TrajGDM & - & - & - & \underline{0.4103} & \underline{0.4875} & \underline{0.6434} & - & - & - & \underline{0.5295} & \underline{0.6302} & \underline{0.7674} \\
UniMob-v1 & 17.93 & 14.01 & 28.65 & 0.4205 & 0.5024 & 0.6570 & 12.70 & 11.65 & 21.94 & 0.5403 & 0.6412 & 0.7889 \\
UniMob-v2 & 17.89 & 13.98 & 28.60 & 0.4228 & 0.5057 & 0.6615 & 12.52 & 11.50 & 21.57 & 0.5439 & 0.6450 & 0.7924 \\
UniMob-v3 & 17.90 & 13.96 & 28.63 & 0.4213 & 0.5040 & 0.6593 & 12.61 & 11.59 & 21.73 & 0.5415 & 0.6436 & 0.7907 \\
UniMob-v4 & \textbf{17.76} & \textbf{13.93} & \textbf{28.50} & \textbf{0.4267} & \textbf{0.5091} & \textbf{0.6653} & \textbf{12.08} & \textbf{11.12} & \textbf{21.03} & \textbf{0.5486} & \textbf{0.6515} & \textbf{0.7993} \\
\bottomrule
\end{tabular}
}
\label{tab:two datasets}
\end{table*}

\begin{table*}[t]
\small
\centering
\caption{Ablation study on Shanghai datasets.}
\vspace{-0.3cm}
\scalebox{1.}{
\begin{tabular}{lcccccc}
\toprule
& \multicolumn{3}{c}{\textbf{Trajectory Prediction}} & \multicolumn{3}{c}{\textbf{Flow Prediction}} \\
\cmidrule(lr){2-4} \cmidrule(lr){5-7}
& \textbf{Acc@1} & \textbf{Acc@3} & \textbf{Acc@5}
& \textbf{MAE} & \textbf{MAPE(\%)} & \textbf{RMSE}\\
\midrule
Ours & 0.4205 & 0.5024 &  0.6570 & 17.93 & 14.01 & 28.65 \\
w/o I2C loss & 0.4165 (-0.95\%) & 0.4906 (-2.35\%) &  0.6487 (-1.26\%) & 18.48 (-2.98\%) & 14.76 (-5.35\%) & 29.91 (-4.21\%) \\
w/o C2I loss & 0.4053 (-3.61\%) & 0.4849 (-3.48\%) &  0.6442 (-1.95\%)
 & 18.27 (-1.86\%) & 14.41 (-2.78\%) & 29.25 (-2.05\%)\\
w/o shared transformer & 0.4115 (-2.14\%) & 0.4882 (-2.83\%) & 0.6461 (-1.66\%) & 18.40 (-2.55\%)
 & 14.60 (-4.21\%) & 29.67 (-3.44\%) \\
w/o flow data & 0.4036(-4.02\%) & 0.4840(-3.66\%) & 0.6421(-2.27\%) & - & - & - \\
w/o trajectory data & - & - & - & 18.56(-3.51\%) & 14.86(-6.07\%) & 30.02(-4.78\%) \\
\bottomrule
\end{tabular}
}
\label{tab:Ablation1}
\end{table*}

\begin{table*}[t]
\small
\centering
\caption{Ablation study on Senegal datasets.}
\vspace{-0.3cm}
\scalebox{1.}{
\begin{tabular}{lcccccc}
\toprule
& \multicolumn{3}{c}{\textbf{Trajectory Prediction}} & \multicolumn{3}{c}{\textbf{Flow Prediction}} \\
\cmidrule(lr){2-4} \cmidrule(lr){5-7}
& \textbf{Acc@1} & \textbf{Acc@3} & \textbf{Acc@5}
& \textbf{MAE} & \textbf{MAPE(\%)} & \textbf{RMSE}\\
\midrule
Ours & 0.5403 & 0.6412 &  0.7889 & 12.70 & 11.65 & 21.94 \\
w/o I2C loss & 0.5371 (-0.59\%)
 & 0.6356 (-0.87\%)
 & 0.7620 (-3.41\%)
 & 13.32 (-4.88\%)
 & 12.12 (-4.03\%)
 & 22.91 (-4.42\%) \\
w/o C2I loss & 0.5285(-2.18\%)
 & 0.6327 (-1.33\%)
 & 0.7476 (-5.24\%)
 & 12.89 (-1.50\%)
 & 11.73 (-0.69\%)
 & 22.10 (-0.73\%)
 \\
w/o shared transformer & 0.5314 (-1.65\%)
 & 0.6331 (-1.26\%)
 & 0.7538 (-4.45\%)
 & 13.20 (-3.94\%)
 & 11.97 (-2.75\%)
 & 22.62 (-3.10\%)
 \\
w/o flow data & 0.5262(-2.61\%) & 0.6297(-1.80\%)
 & 0.7548(-4.32\%) & - & - & - \\
w/o trajectory data & - & - & - & 13.40(-5.51\%)
 & 12.18(-4.55\%) & 22.98(-4.74\%) \\
\bottomrule
\end{tabular}
}
\label{tab:Ablation3}
\end{table*}

\subsection{Overall Performance}
As shown in Tables~\ref{tab:two datasets}, our method demonstrates similar or better performance than the state-of-the-art baselines for all tasks on Shanghai and Senegal datasets (Please refer to Table~\ref{tab:Xinjiang} in Appendix~\ref{sec::Overall Performance} for Xingjiang dataset). We conducted multiple experiments and reported the average performance.
In flow and trajectory prediction tasks conducted on multiple real-world datasets, our UniMob model demonstrated the best performance across all evaluation metrics. Specifically, it achieved a performance improvement of over 6\% in flow prediction and 3.73\% increase in trajectory prediction.
Additionally, compared to other baseline methods, only our model can simultaneously perform flow and trajectory predictions, demonstrating that our model design effectively achieves unified human mobility prediction.
Furthermore, we used four model variants for each task. Each variant outperformed other baseline methods, maintaining flexibility to handle different scenarios while demonstrating excellent performance.
The above conclusions fully demonstrate the feasibility of a unified model in human mobility prediction. Our UniMob model can handle various types of mobility data, showcasing exceptional scalability and robustness. As the first attempt to propose a universal model paradigm for mobility prediction, we have successfully expanded the boundaries of this field.

Notably, mobility prediction models based on diffusion models, such as PriSTI and TrajGDM, demonstrate superior performance compared to other baselines. This underscores the powerful modeling capability of diffusion models in capturing the spatiotemporal correlations of mobility data. Diffusion models effectively handle dynamics and uncertainties in mobility data through an iterative denoising process, significantly enhancing prediction performance. Therefore, our UniMob model leverages diffusion models to accurately capture spatiotemporal dependencies in mobility data accurately, proving its effectiveness.

\subsection{Ablation Study}
To evaluate the impact of each module in UniMob, we conducted ablation experiments, divided into ablations of model design and data usage.
\textbf{Model Design:}
(1) w/o I2C loss: This variant keeps the model structure unchanged but removes the I2C loss.
(2) w/o C2I loss: Similar to the previous one, this variant only removes the C2I loss.
(3) w/o shared transformer: In this variant, the flow and trajectory losses no longer share a transformer; instead, each has its independent transformer.
\textbf{Data Usage:}
(4) w/o flow data: The model is trained using only trajectory data.
(5) w/o trajectory data: The model is trained using only flow data.

The results of the ablation experiments conducted on the Shanghai and Senegal datasets are shown in Tables~\ref{tab:Ablation1} and ~\ref{tab:Ablation3} (see Table~\ref{tab:Ablation2} in Appendix~\ref{sec::ablation} for the Xinjiang dataset). For the ablation experiments on model design, it is evident that the shared transformer offers limited benefits for interacting with different mobility data types. The most significant performance improvements come from task-specific loss functions. For instance, the I2C loss enhances flow prediction by using aggregated trajectory and flow data for spatiotemporal alignment. Similarly, the C2I loss uses contrastive learning to construct positive samples of flow and trajectory with similar spatiotemporal patterns, thereby aligning macro and micro mobility distribution. These experiments highlight the effectiveness of our approach in aligning trajectory and flow data.

We removed different data types for the ablation experiments on data usage and trained the model using only a single type of mobility data. The results showed a significant performance decline. This demonstrates the effectiveness and importance of our model in utilizing different types of mobility data. By combining multiple data types, UniMob can more comprehensively understand and predict human mobility behavior, thereby significantly enhancing the model's overall performance.

\subsection{Noise Perturbation}
In real life, mobility data often contains noise. This noise can arise from various sources, such as errors produced by sensors during the collection process or intentionally added by data operators to protect user privacy.  To assess our UniMob model's robustness, we added noise to the data and evaluated its performance.

\begin{figure}[t]
\centering
\subfigure[Shanghai]{\includegraphics[width=.23\textwidth]{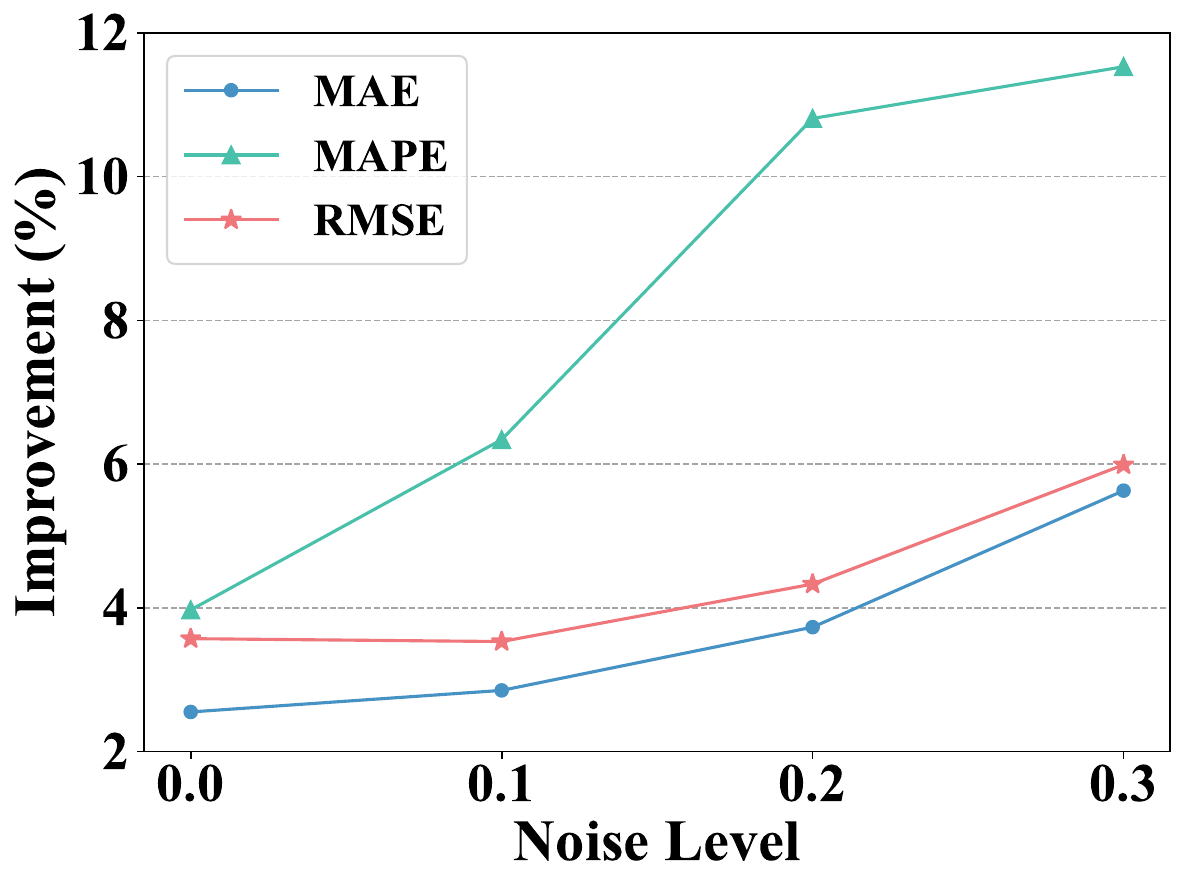}}
\vspace{-0.3cm}
\subfigure[Senegal]{\includegraphics[width=.23\textwidth]{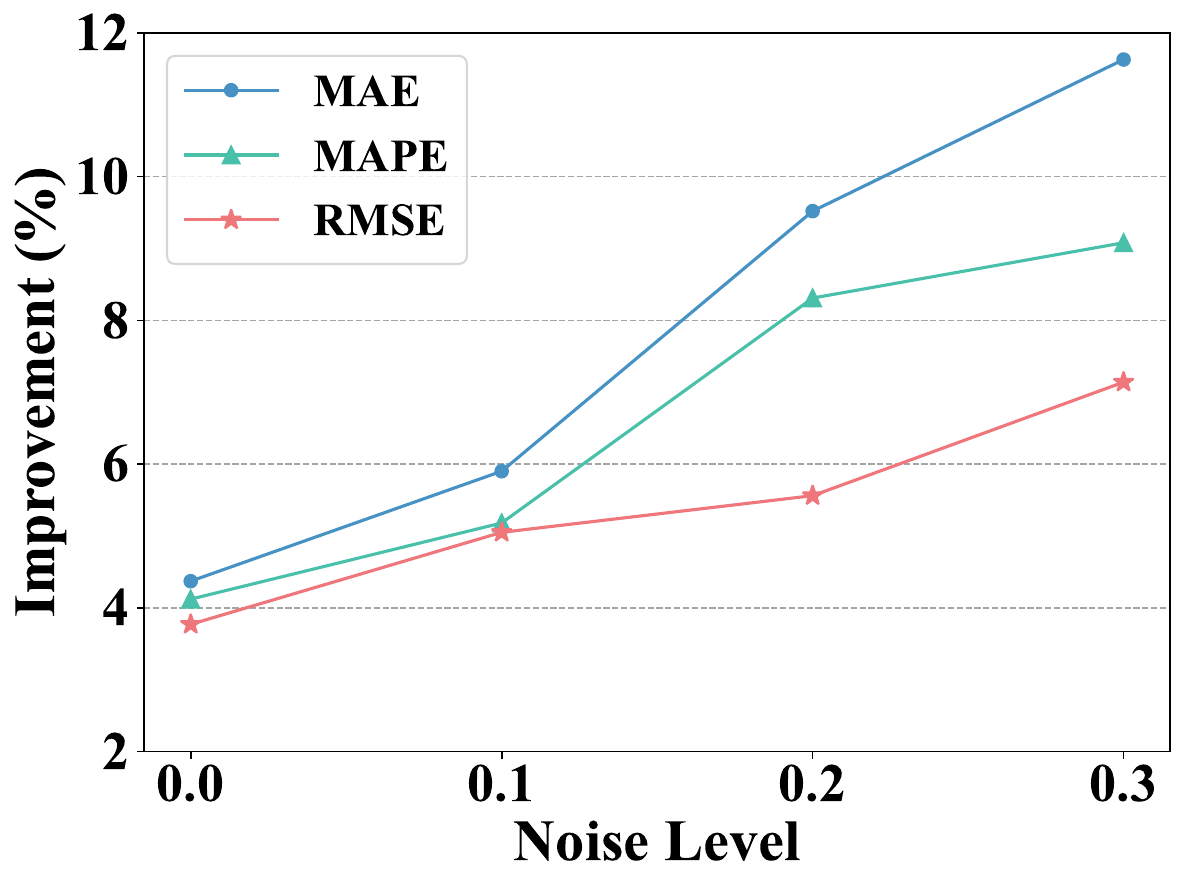}}
\caption{Flow prediction with noisy data on Shanghai and Senegal datasets.} 
\vspace{-0.3cm}
\label{fig:noisy_flow}
\end{figure}

\begin{figure}[t]
\centering
\subfigure[Shanghai]{\includegraphics[width=.23\textwidth]{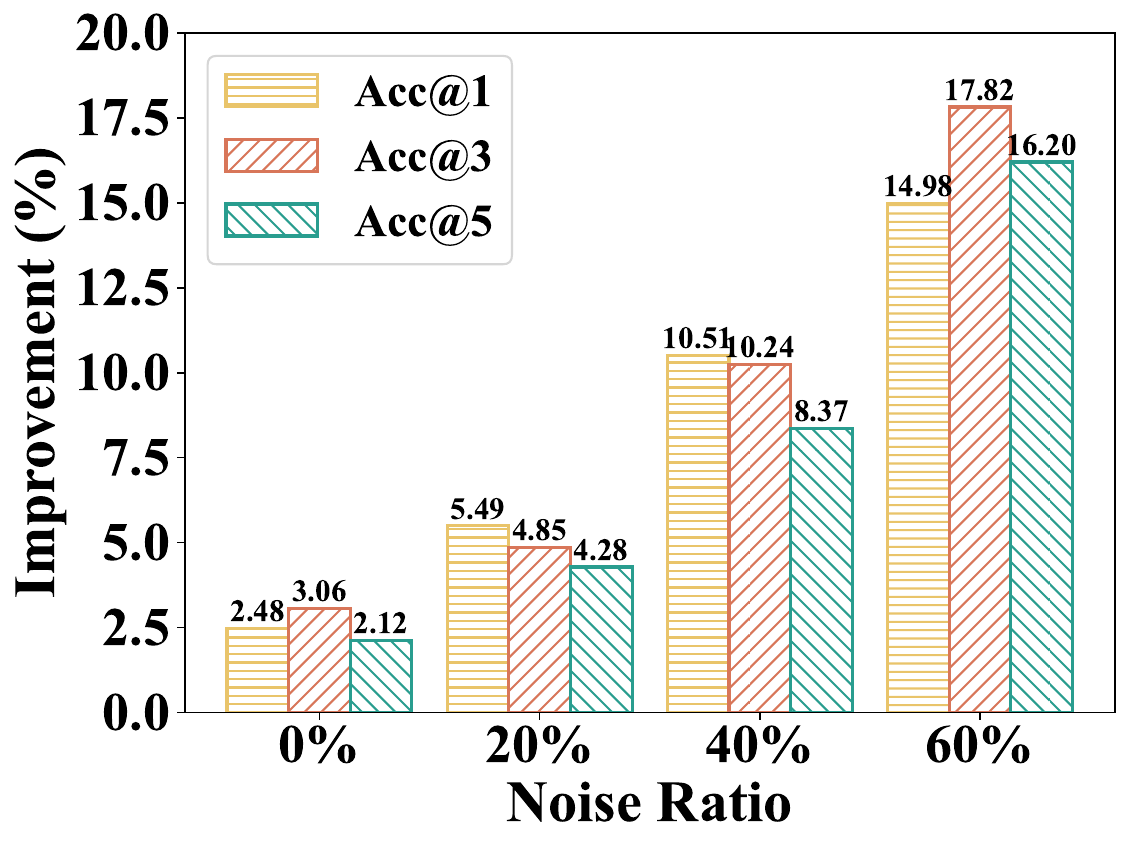}}
\vspace{-0.3cm}
\subfigure[Senegal]{\includegraphics[width=.23\textwidth]{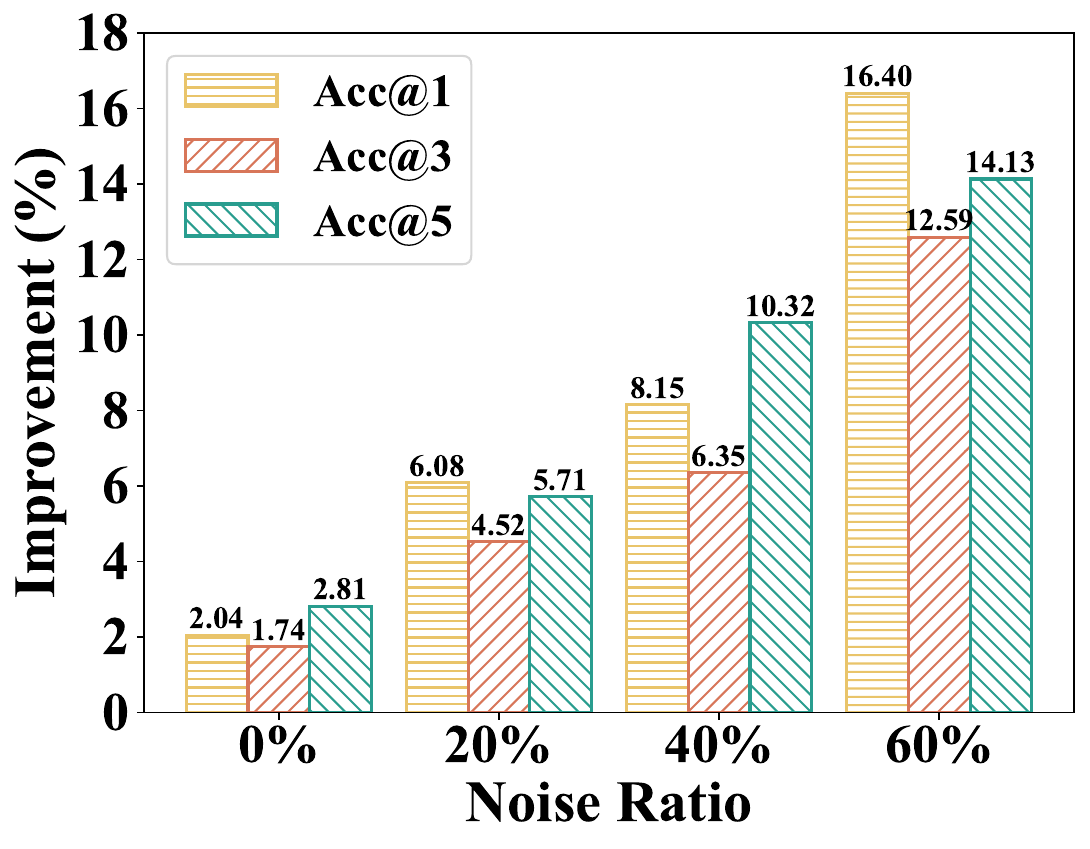}}
\caption{Trajectory prediction with noisy data on Shanghai and Senegal datasets.} 
\vspace{-0.3cm}
\label{fig:noisy_trajectory}
\end{figure}

For flow data, we introduced varying noise levels to simulate different degrees of data quality. Figure~\ref{fig:noisy_flow} shows that our model's improvement over the best baseline is relatively small without noise. When the noise level reaches 0.3, our model demonstrates a relative improvement of more than 10\%. This indicates that compared to other baseline models, UniMob exhibits better robustness in handling noisy data, making it more capable of adapting to flow data with noise for prediction. Moreover, we experimented with adding different noise levels to trajectories. Figure~\ref{fig:noisy_trajectory} shows that as the noise ratio increases, the improvement of our model relative to the best baseline also increases, achieving a maximum gain of up to 17.82\%. Because UniMob integrates two types of mobility data, allowing one type of data to provide the same spatiotemporal dynamics as a supplement when the other type of data is noisy, thereby enhancing the model's robustness. The synergistic effects between different data types can still provide reliable predictions even in noisy data.

\subsection{Few-shot Performance}
Similarly, the amount of mobility data may be limited in real-world scenarios due to privacy concerns, data collection challenges, or other constraints. To simulate this situation, we reduce the amount of flow and trajectory data through different operations.

\begin{figure}[t]
\centering
\subfigure[Shanghai]{\includegraphics[width=.23\textwidth]{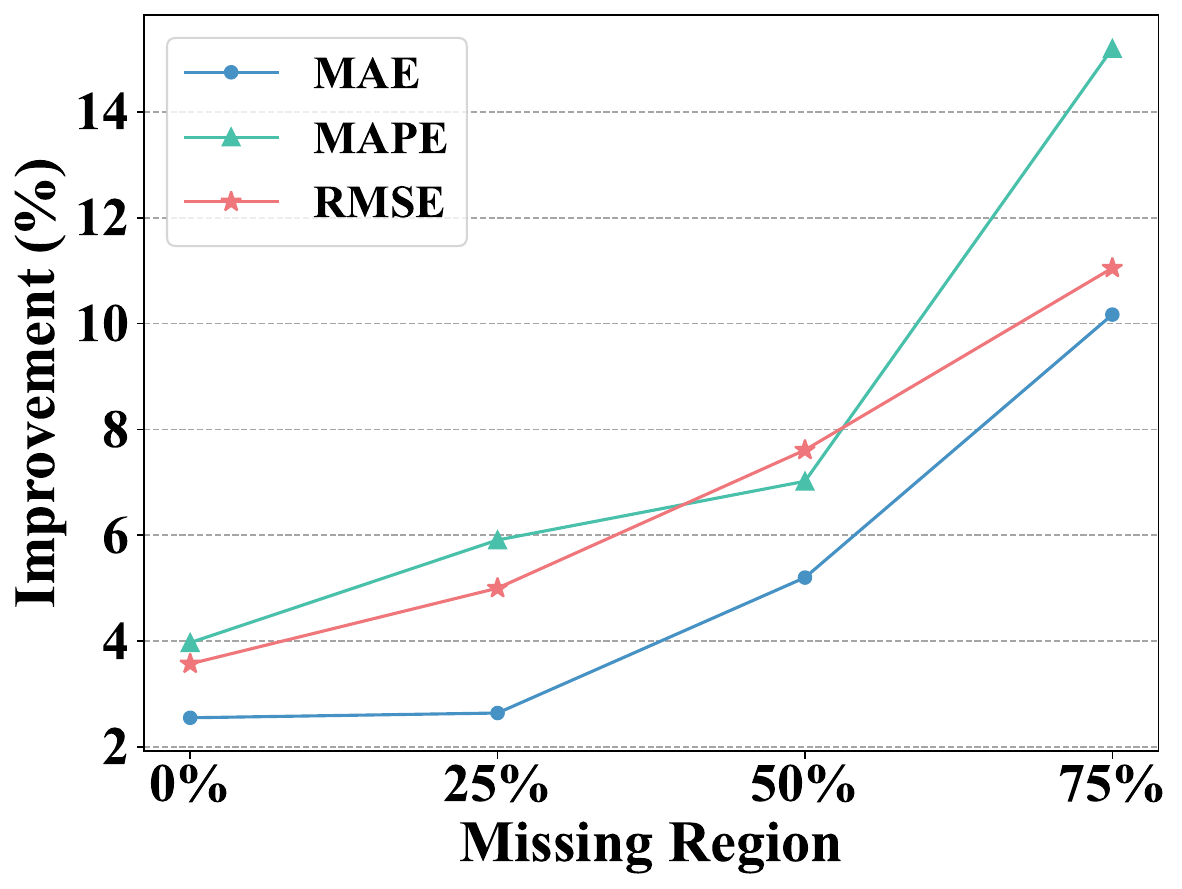}}
\vspace{-0.3cm}
\subfigure[Senegal]{\includegraphics[width=.23\textwidth]{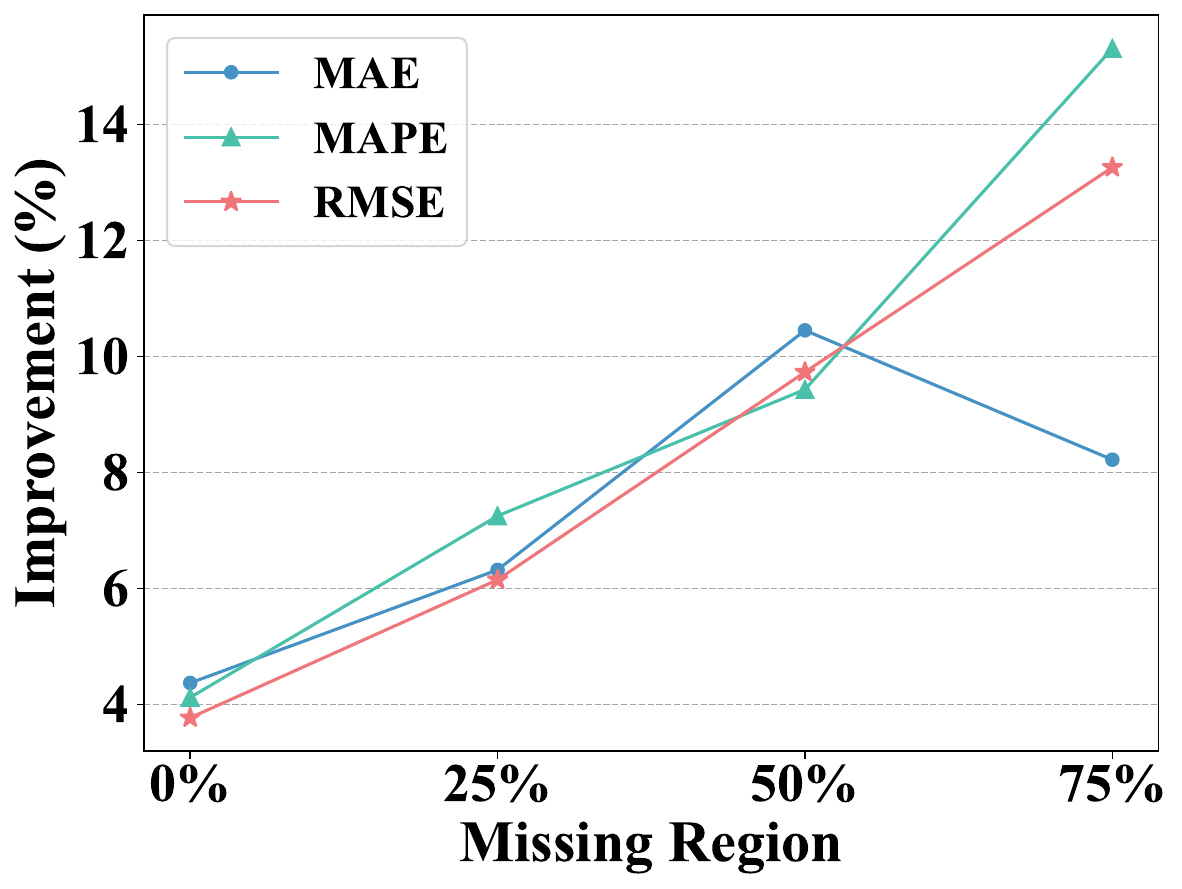}}
\caption{Flow prediction with scarce data on Shanghai and Senegal datasets.} 
\vspace{-0.3cm}
\label{fig:low_flow}
\end{figure}

\begin{figure}[t]
\centering
\subfigure[Shanghai]{\includegraphics[width=.23\textwidth]{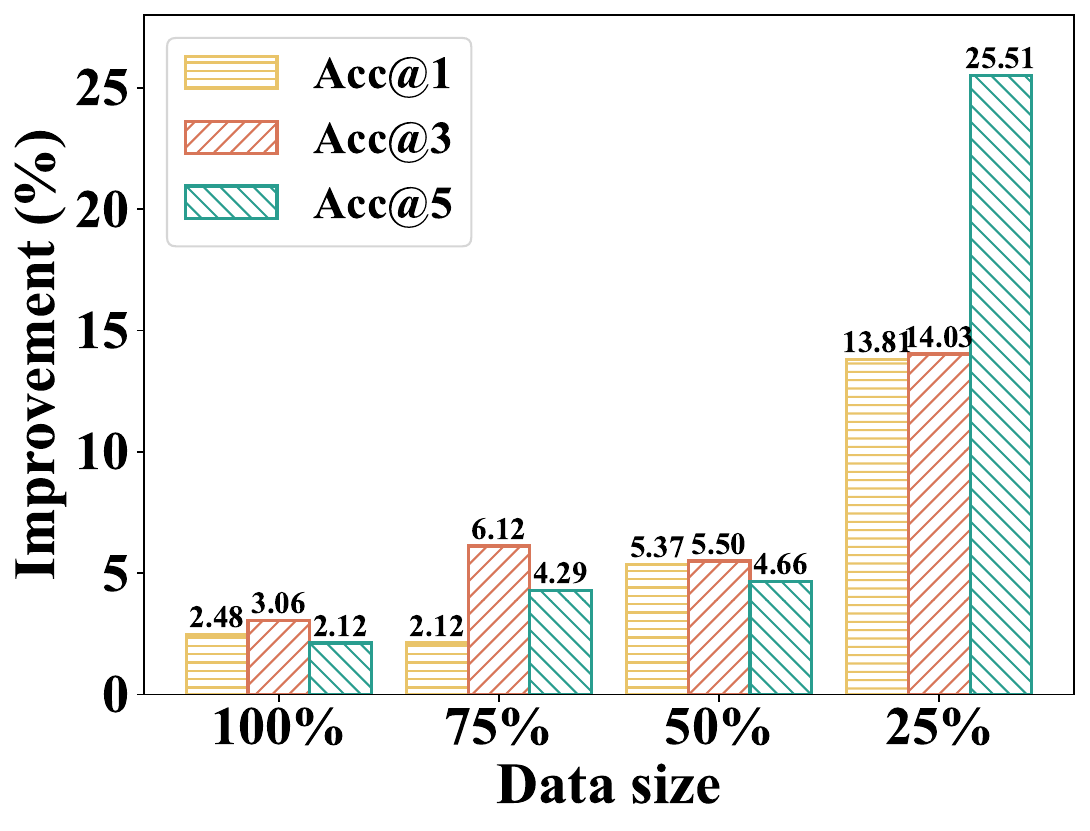}}
\vspace{-0.3cm}
\subfigure[Senegal]{\includegraphics[width=.23\textwidth]{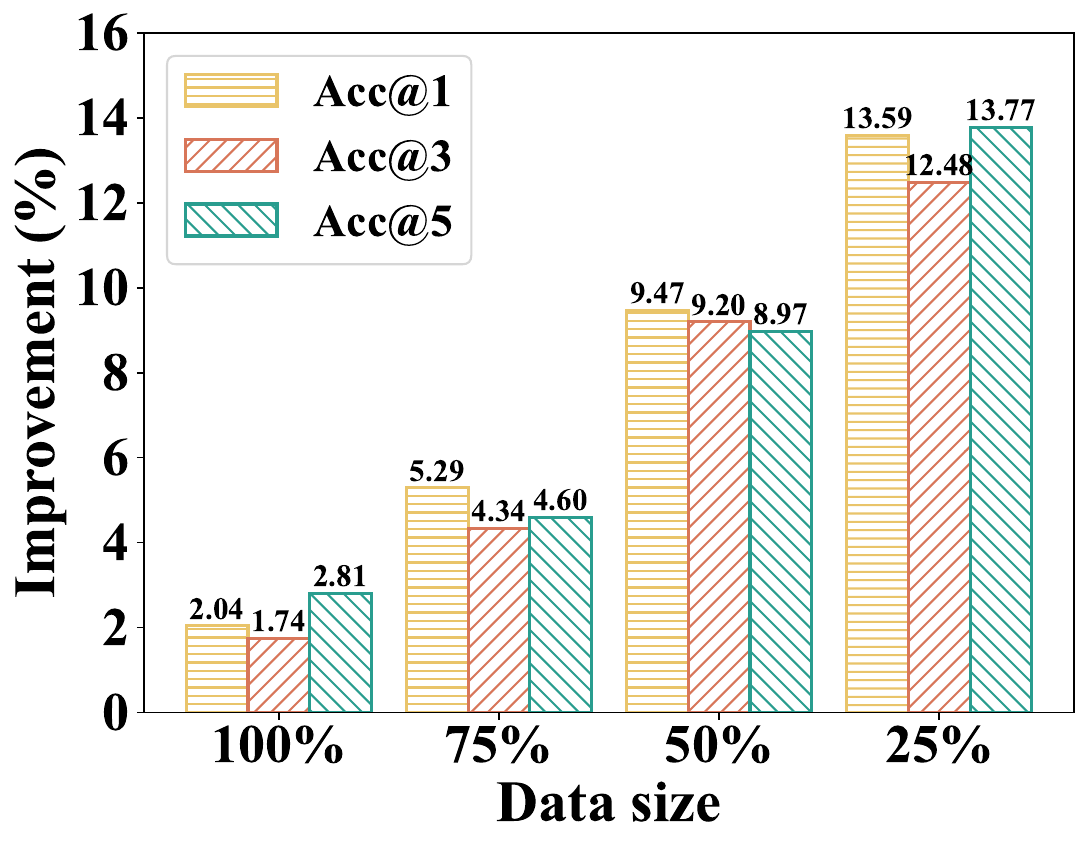}}
\caption{Trajectory prediction with scarce data on Shanghai and Senegal datasets.} 
\vspace{-0.3cm}
\label{fig:low_trajectory}
\end{figure}

As shown in Figure~\ref{fig:low_flow}, we constructed scenarios with varying proportions of locations having missing flow records. As the proportion of regions with missing flow data increased, our model still demonstrated a significant performance improvement compared to the best baseline. For instance, in the Shanghai dataset, UniMob achieves an improvement of up to 14\% when 75\% of the region is missing. UniMob's robustness is evident in its ability to maintain high performance despite the absence of a substantial amount of flow data. This is due to its ability to leverage the available trajectory data, compensating for the missing flow information through its joint modeling approach. 
As shown in Figure~\ref{fig:low_trajectory}, we used datasets of different sizes for trajectory data to explore the performance of trajectory prediction with limited data. When the amount of trajectory data is very limited (e.g., only 25\% of the dataset), our model shows a 25\% improvement in the Shanghai dataset compared to the best baseline. This indicates that when trajectory data is scarce, the flow data provides more diverse mobility patterns, effectively compensating for the lack of trajectory data.

By effectively utilizing the spatiotemporal correlations between different types of mobility data, UniMob can provide accurate predictions even in data-scarce environments. UniMob's ability to deliver reliable predictions with limited data highlights its robustness and practical applicability in various scenarios, ensuring dependable performance regardless of data constraints.

\section{Conclusion}
In this paper, we address an important problem of unified human mobility prediction.
We propose a universal human mobility prediction model named UniMob, achieving broad adaptability to various data formats and characteristics. 
UniMob successfully captures the spatiotemporal dynamics inherent in different modalities of mobility data by unified tokenization and bidirectional alignment between them, enabling unified modeling.
Extensive experiments on real-world datasets demonstrate that UniMob outperforms in trajectory and flow prediction. Moreover, UniMob can flexibly adapt to noisy and scarce data scenarios, showcasing its robustness.
In the future, we will explore integrating additional urban data modalities, such as weather data, social network data, and GIS data. These factors influence human mobility, we can better predict human mobility patterns by combining them with mobility data.


\clearpage
\bibliographystyle{ACM-Reference-Format}
\balance
\bibliography{sample-base}
\clearpage

\appendix
\balance

\section{APPENDIX FOR REPRODUCIBILITY}
\subsection{Related Work}
\subsubsection{Diffusion Models}\label{sec::diffusion}
The diffusion model is a probabilistic generative model first introduced by Sohl-Dickstein et al.~\cite{sohl2015deep} and further improved by Ho et al. ~\cite{ho2020denoising} and Song et al. ~\cite{song2020score}. As a novel generative model, diffusion models have rapidly advanced in time series and spatio-temporal modeling. 
Research on time series modeling based on diffusion models is widely applied, such as time series imputation ~\cite{alcaraz2022diffusion,liu2023pristi}, time series generation~\cite{lim2023regular,lin2023diffusion}, and time series forecasting~\cite{li2022generative,bilovs2022modeling}. 
DiffSTG~\cite{wen2023diffstg} is the first attempt to generalize the widespread denoising diffusion probabilistic models to spatiotemporal graphs (STGs), leading to a novel non-autoregressive framework. 
KSTDiff~\cite{zhou2023towards} designed a knowledge-enhanced denoising network to capture the spatiotemporal dependencies of urban flows and the influence of the urban environment in the denoising process.
DiffTraj~\cite{zhou2023towards} is a spatiotemporal diffusion probabilistic model for trajectory generation. This model effectively combines the generative capabilities of diffusion models with spatiotemporal features derived from real trajectories.
In this work, we introduce the diffusion model for unified mobility prediction adapted to different data types.

\subsection{Datasets Details}\label{sec::datasets_info}
We conducted extensive experiments on three real-world mobility datasets: Shanghai, Senegal, and Xinjiang. The details of datasets are summarized in Table~\ref{table:datasets}. We preprocess the trajectory data for three datasets, filtering out users with fewer than five records per day. For location preprocessing, we map GPS points to predefined grid IDs of a specific granularity. For temporal preprocessing, we organize the time data into fixed intervals, such as hourly or half-hourly segments. Finally, we divide the data into training, validation, and testing sets in a 7:1:2 ratio in chronological order. 

\begin{table}[h]
\setlength{\abovecaptionskip}{0.cm}
\setlength{\belowcaptionskip}{-0.cm}
\caption{Basic statistics of mobility datasets.}
\label{table:datasets}
\begin{center}
\scalebox{0.9}{
\begin{tabular}{ >{\centering\arraybackslash}m{1cm} 
>{\centering\arraybackslash}m{1cm} 
>{\centering\arraybackslash}m{1.2cm} 
>{\centering\arraybackslash}m{1cm}
>{\centering\arraybackslash}m{1cm}
>{\centering\arraybackslash}m{1cm}}
 \hline
City  & Duration & Users & Location \\ 
 \hline
Shanghai & 7 days & 700000 & 4096 \\ 
Senegal & 14 days & 8000 & 1666 \\ 
Xinjiang & 28 days & 1200000 & 4096\\ 
\hline
\end{tabular}}
\end{center}
\vspace{-0.3cm}
\end{table}

\subsection{Baselines}\label{sec::baselines}
To evaluate the performance of our proposed model, we compared it with state-of-the-art models. Previous methods could only accomplish one type of mobility data prediction task, so the baseline methods are divided into trajectory and flow prediction. 

\paragraph{Flow Prediction} The baselines for flow prediction are as follows:
\begin{itemize}[leftmargin=*]
\item \textbf{HA}~\cite{sun2020predicting}: It considers the inflow and outflow to be seasonal processes and employs the average of the previous seasons as the prediction for a week-long period. 
\item \textbf{VAR}~\cite{lu2016integrating}: This method is vector autoregressive single-step predictor.
\item \textbf{ST-ResNet}~\cite{zhang2017deep}: ST-ResNet employs the residual neural network framework to model the temporal closeness, period, and trend properties of crowd flow.
\item \textbf{MSDR}~\cite{liu2022msdr}: Multi-Step Dependency Relationship (MSDR) is a brand new variant of recurrent neural networks. Instead of only looking at the hidden state from the latest time step, MSDR explicitly takes those from multiple historical time steps as the input of each time unit.
\item \textbf{STID}~\cite{shao2022spatial}: A simple multi-layer perceptron addresses the indistinguishability of time series samples in spatial and temporal dimensions.
\item \textbf{PriSTI}~\cite{liu2023pristi}: This method extracts coarse but effective spatiotemporal dependencies from conditional information using a diffusion model, serving as a global context prior.
\end{itemize}

\paragraph{Trajectory Prediction} The baselines for trajectory prediction are as follows:
\begin{itemize}[leftmargin=*]
\item \textbf{Markov Model}~\cite{gambs2012next}: The Markov model is a statistical model used to describe the change of states over time. It uses historical trajectory data for location prediction by calculating the transition probabilities between these locations.
\item \textbf{LSTM}~\cite{Kong2018HST}: The LSTM network is good at handling sequential data and has the advantage of encoding long-term dependencies, which can naturally be applied to location prediction.
\item \textbf{DeepMove}~\cite{feng2018deepmove}: The method designs a multimodal embedding recurrent neural network to capture complex sequential transitions by jointly embedding multiple factors that control human mobility.
\item \textbf{STAN}~\cite{luo2021stan}: This model associates non-contiguous but functionally similar visited points that are not adjacent to each other to predict the next location.
\item \textbf{SNPM}~\cite{yin2023next}: The method constructs a Sequence-based, Dynamic Neighbor Graph (SDNG) to find the similarity neighborhood and develop a Multi-Step Dependency Prediction model.
\item \textbf{TrajGDM}~\cite{chu2024simulating}: The method utilizes diffusion models to capture the universal mobility pattern in a trajectory dataset for trajectory prediction.
\item \textbf{GETNext}~\cite{yang2022getnext}: The method employs a global trajectory flow map and a novel Graph Enhanced Transformer model to leverage collaborative signals for more accurate trajectory prediction.
\end{itemize}

\begin{table}[h]
\small
\centering
\caption{Overall Performance on Xinjiang datasets.}
\vspace{-0.3cm}
\scalebox{0.9}{
\begin{tabular}{lcccccc}
\toprule
& \multicolumn{3}{c}{\textbf{Flow Prediction}} & \multicolumn{3}{c}{\textbf{Trajectory Prediction}} \\
\cmidrule(lr){2-4} \cmidrule(lr){5-7}
& \textbf{MAE} & \textbf{MAPE(\%)} & \textbf{RMSE}
& \textbf{Acc@1} & \textbf{Acc@3} & \textbf{Acc@5}\\
\midrule
HA & 33.16 & 30.54 & 44.28 & - & - & - \\
VAR & 23.90 & 22.15 & 36.63 & - & - & - \\
ST-ResNet & 19.72 & 17.36 & 31.56 & - & - & - \\
MSDR & 17.95 & 16.53 & 29.60 & - & - & - \\
STID & 17.01 & 15.70 & 27.36 & - & - & - \\
PriSTI & \underline{16.80} & \underline{15.37} & \underline{26.47} & - & - & - \\
Markov & - & - & - & 0.3156 & 0.3924 & 0.4571 \\
LSTM & - & - & - & 0.3847 & 0.4519 & 0.5450 \\
DeepMove & - & - & - & 0.4261 & 0.5143 & 0.6318 \\
STAN & - & - & - & 0.4432 & 0.5307 & 0.6609 \\
SNPM & - & - & - & 0.4618 & 0.5574 & 0.6926 \\
GETNext & - & - & - & 0.4650 & 0.5598 & 0.6975 \\
TrajGDM & - & - & - & \underline{0.4673} & \underline{0.5632} & \underline{0.7054} \\
UniMob-v1 & 16.31 & 14.91 & 25.98 & 0.4768 & 0.5795 & 0.7217 \\
UniMob-v2 & 15.96 & 14.72 & 25.54 & 0.4815 & 0.5853 & 0.7286 \\
UniMob-v3 & 16.12 & 14.84 & 25.70 & 0.4791 & 0.5830 & 0.7253 \\
UniMob-v4 & \bf{15.87} & \bf{14.50} & \bf{25.19} & \bf{0.4841} & \bf{0.5897} & \bf{0.7336} \\
\bottomrule
\end{tabular}
}
\label{tab:Xinjiang}
\end{table}

\begin{table*}[t]
\small
\centering
\caption{Ablation study on Xinjiang datasets.}
\vspace{-0.3cm}
\scalebox{1.}{
\begin{tabular}{lcccccc}
\toprule
& \multicolumn{3}{c}{\textbf{Trajectory Prediction}} & \multicolumn{3}{c}{\textbf{Flow Prediction}} \\
\cmidrule(lr){2-4} \cmidrule(lr){5-7}
& \textbf{Acc@1} & \textbf{Acc@3} & \textbf{Acc@5}
& \textbf{MAE} & \textbf{MAPE(\%)} & \textbf{RMSE}\\
\midrule
Ours & 0.4768 & 0.5795 &  0.7217 & 16.31 & 14.91 & 25.98 \\
w/o I2C loss & 0.4736 (-0.67\%) & 0.5730 (-1.12\%) &  0.7125 (-1.28\%)
 & 16.78 (-2.88\%) & 15.43 (-3.49\%) & 27.02 (-4.00\%) \\
w/o C2I loss & 0.4689 (-1.66\%) & 0.5671 (-2.14\%) &  0.7064 (-2.12\%)
 & 16.46 (-0.92\%) & 15.08 (-1.14\%) & 26.90 (-3.54\%) \\
w/o shared transformer & 0.4702 (-1.39\%) & 0.5693 (-1.76\%)
 & 0.7091 (-1.75\%) & 16.67 (-2.21\%) & 15.29 (-2.55\%)
 & 26.97 (-3.81\%)
 \\
w/o flow data & 0.4639(-2.71\%) & 0.5620(-3.02\%)
 & 0.6998(-3.03\%) & - & - & - \\
w/o trajectory data & - & - & - & 16.87(-3.43\%)
 & 15.56(-4.36\%) & 27.20(-4.70\%) \\
\bottomrule
\end{tabular}
}
\label{tab:Ablation2}
\end{table*}

\section{Experimental Performance}\label{sec::Results}
\subsection{Overall Performance}\label{sec::Overall Performance}
Table~\ref{tab:Xinjiang} shows the performance of our universal mobility prediction model on the Xinjiang dataset. UniMob not only accomplishes both trajectory and flow predictions simultaneously but also surpasses current advanced baseline models in all evaluation metrics. Specifically, it achieves 5.34\% performance improvement in flow prediction and more than 4\% enhancement in trajectory prediction. These results fully demonstrate the generality and reliability of our model.

\subsection{Ablation study}\label{sec::ablation}
We conducted ablation experiments on two aspects: model design and data utilization. By sequentially removing components of the model design, we identified three design elements that align with different data formats and distributions, each impacting performance, thus validating their effectiveness. Regarding data utilization, by replacing multi-type data with single-type data, we visually demonstrated the performance enhancement brought by using multi-type mobility data in human mobility prediction through our universal model.

\subsection{Noise Perturbation}\label{sec::noise}

\begin{figure}[t]
\centering
\subfigure[Flow prediction]{\includegraphics[width=.23\textwidth]{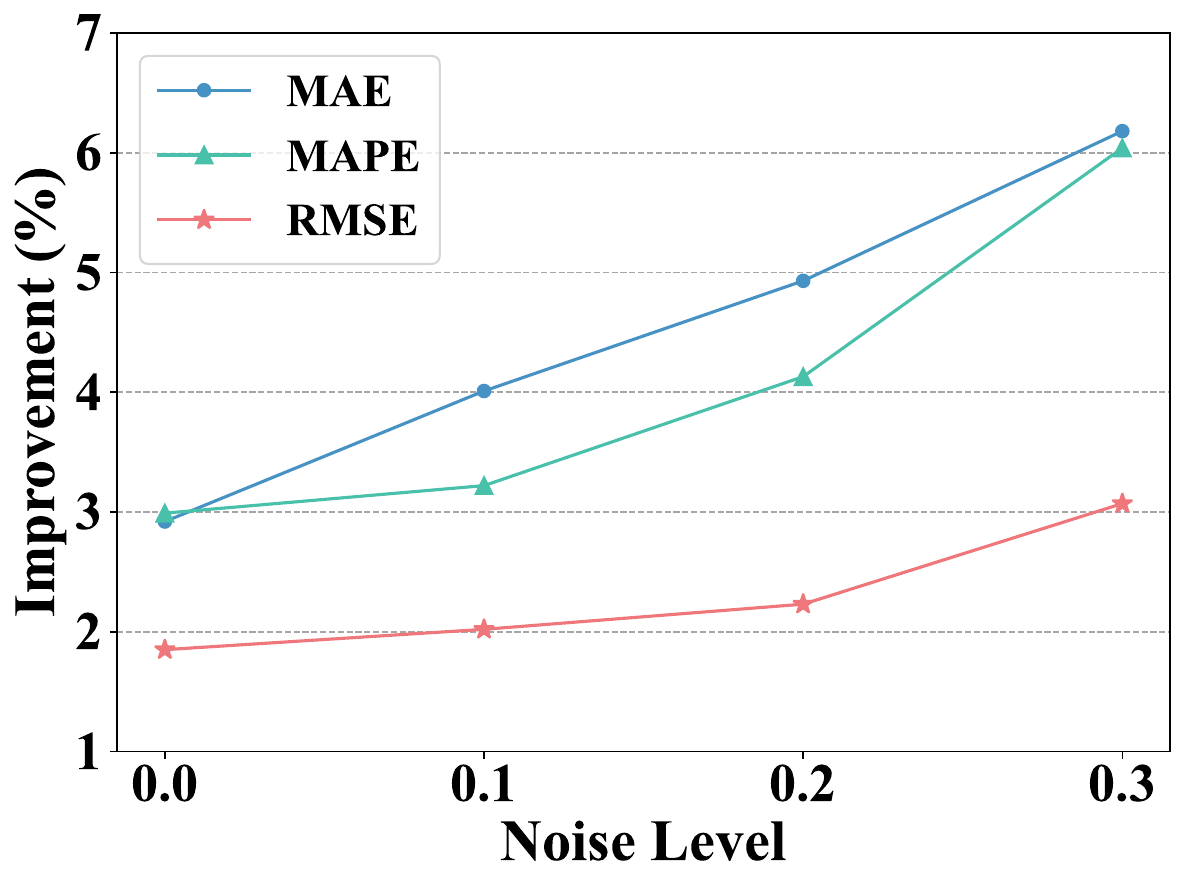}}
\vspace{-0.5cm}
\subfigure[Trajectory prediction]{\includegraphics[width=.23\textwidth]{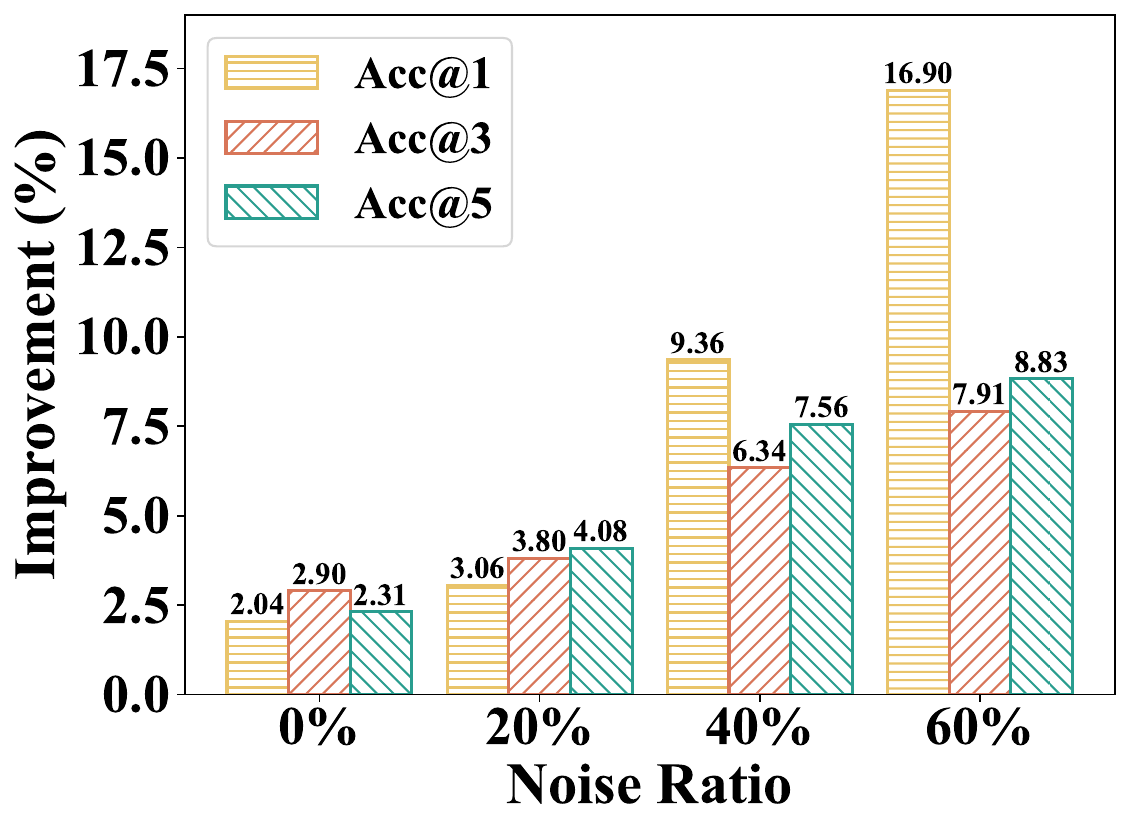}}
\caption{Flow and trajectory prediction with noisy data on Xinjiang dataset.} 
\label{fig:noisy}
\end{figure}

Due to biases from sensor collection and artificial noise added for privacy protection, the data used for mobility prediction often contains noise. To verify whether our model can still maintain good predictive capabilities in noisy conditions, we added noise to both the flow and trajectory data. Figure~\ref{fig:noisy} shows that as noise levels increase, our model continues to outperform the best baseline model, and our performance advantage becomes even more pronounced relative to the baseline with increasing noise. This effectively demonstrates the high robustness of our UniMob model.

\subsection{Few-shot Performance}\label{sec::few-shot}
\begin{figure}[t]
\centering
\subfigure[Flow prediction]{\includegraphics[width=.23\textwidth]{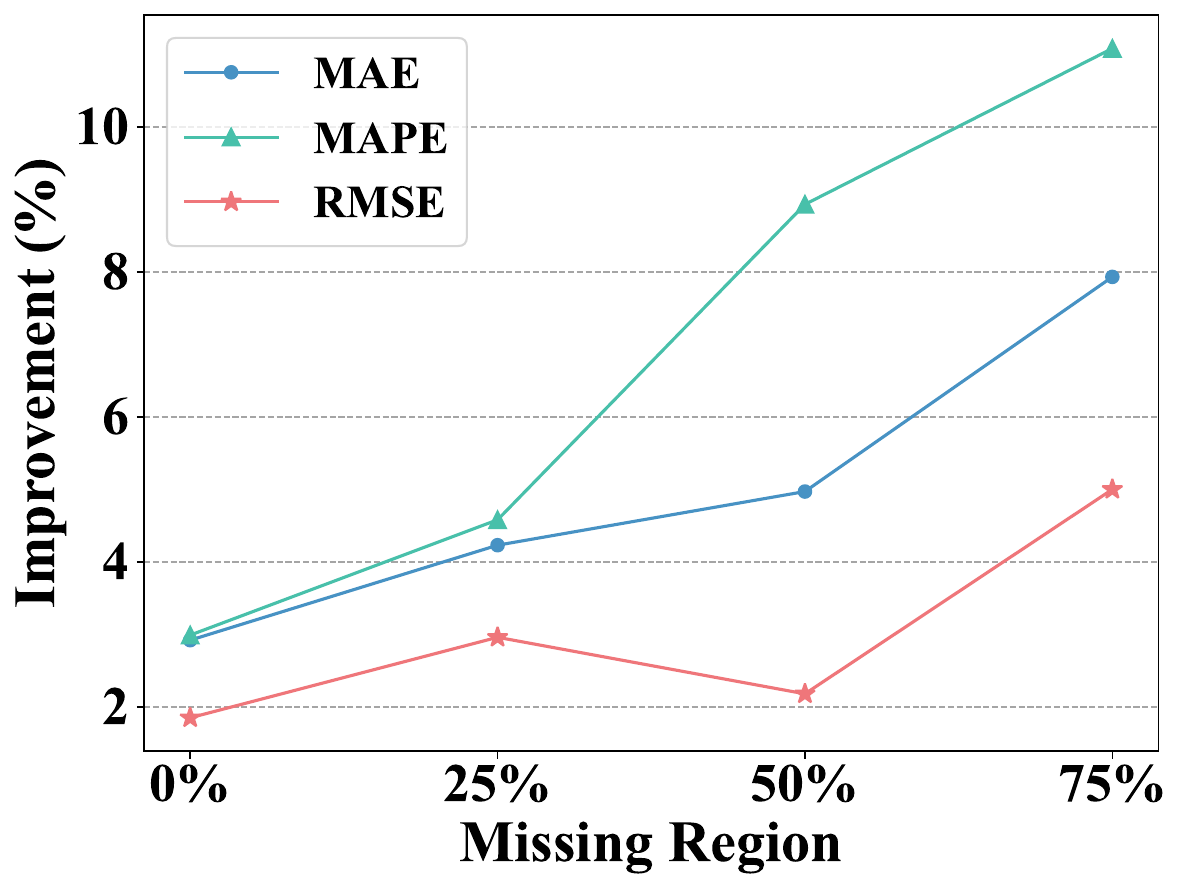}}
\vspace{-0.5cm}
\subfigure[Trajectory prediction]{\includegraphics[width=.23\textwidth]{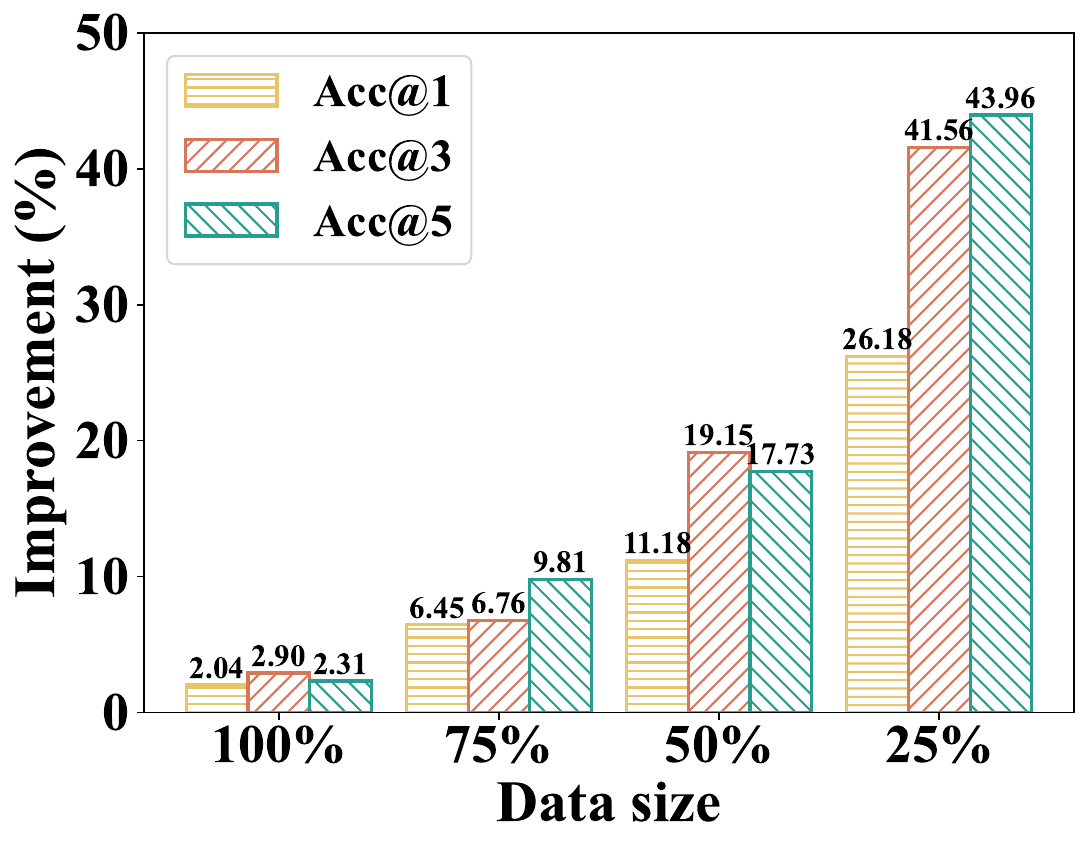}}
\caption{Flow and trajectory prediction with scarce data on Xinjiang dataset.} 
\label{fig:low}
\end{figure}

Similarly, due to data collection and privacy protection limitations, the amount of mobility data we acquire is often limited. Therefore, we tested the few-shot learning capabilities of our UniMob model. As shown in Figure~\ref{fig:low}, our model still performs excellently even in a data-constrained environment.

\end{document}